%% file: aaai2026.tex
\documentclass[letterpaper]{article} % DO NOT CHANGE THIS
\usepackage{aaai2026}  % DO NOT CHANGE THIS
\usepackage{times}  % DO NOT CHANGE THIS
\usepackage{helvet}  % DO NOT CHANGE THIS
\usepackage{courier}  % DO NOT CHANGE THIS
\usepackage[hyphens]{url}  % DO NOT CHANGE THIS
\usepackage{graphicx} % DO NOT CHANGE THIS
\usepackage{natbib}  % DO NOT CHANGE THIS AND DO NOT ADD ANY OPTIONS TO IT
\usepackage{caption} % DO NOT CHANGE THIS AND DO NOT ADD ANY OPTIONS TO IT
\usepackage{algorithm}
\usepackage{algorithmic}
\usepackage{amsmath} 
\usepackage{amssymb} 
\usepackage{listings}
\usepackage{cleveref}

\DeclareCaptionStyle{ruled}{labelfont=normalfont,labelsep=colon,strut=off} % DO NOT CHANGE THIS
\floatstyle{ruled}
\newfloat{listing}{tb}{lst}{}
\floatname{listing}{Listing}
\usepackage{pifont}
\newcommand{\nomark}{\ding{55}}   % ✗

\usepackage{multirow}
\usepackage{booktabs}
\usepackage{graphicx}
\usepackage{subcaption}

\title{PlantTraitNet: An Uncertainty-Aware Multimodal Framework for Global-Scale Plant Trait Inference from Citizen Science Data}
\author{
    Ayushi Sharma\textsuperscript{\rm 1}, Johanna Trost\textsuperscript{\rm 1}, Daniel Lusk\textsuperscript{\rm 1}, Johannes Dollinger\textsuperscript{\rm 2}, Julian Schrader\textsuperscript{\rm 3}, Christian Rossi\textsuperscript{\rm 4}, Javier Lopatin\textsuperscript{\rm 5}, Etienne Laliberté\textsuperscript{\rm 6}, Simon Haberstroh\textsuperscript{\rm 7}, Jana Eichel\textsuperscript{\rm 8}, Daniel Mederer\textsuperscript{\rm 9}, Jose Miguel Cerda-Paredes\textsuperscript{\rm 10, \rm5}, Shyam S. Phartyal\textsuperscript{\rm 11}, Lisa-Maricia Schwarz\textsuperscript{\rm 12, \rm13}, Anja Linstädter\textsuperscript{\rm 12}, Maria Conceição Caldeira\textsuperscript{\rm 14}, Teja Kattenborn\textsuperscript{\rm 1}
}
\affiliations{
    \textsuperscript{\rm 1}Chair of Sensor-based Geoinformatics, University of Freiburg, Germany\\
    \textsuperscript{\rm 2}EcoVision Lab, DM3L, University of Zurich, Switzerland\\
    \textsuperscript{\rm 3}Department of Biological Sciences, Macquarie University, Australia\\
    \textsuperscript{\rm 4}Swiss National Park, Switzerland\\
    \textsuperscript{\rm 5}Facultad de Ingeniería y Ciencias, Universidad Adolfo Ibáñez, Chile\\
    \textsuperscript{\rm 6}Department of Biological Sciences, Université de Montréal, Canada\\
    \textsuperscript{\rm 7}Chair of Ecosystem Physiology, University of Freiburg, Germany\\
    \textsuperscript{\rm 8}Department of Physical Geography, Utrecht University, The Netherlands\\
    \textsuperscript{\rm 9}Institute for Earth System Science and Remote Sensing, Leipzig University, Germany\\
    \textsuperscript{\rm 10}Data Observatory, Universidad Adolfo Ibáñez, Chile\\
    \textsuperscript{\rm 11}Department of Forestry, Mizoram University, India\\
    \textsuperscript{\rm 12}Biodiversity Research / Systematic Botany, University of Potsdam, Germany\\
    \textsuperscript{\rm 13}Department of Plant Nutrition, Institute of Crop Science and Resource Conservation, University of Bonn, Germany\\
    \textsuperscript{\rm 14}Forest Research Centre, School of Agriculture, University of Lisbon, Portugal\\

    ayushi.sharma@geosense.uni-freiburg.de, teja.kattenborn@geosense.uni-freiburg.de
}

\begin{document}

\maketitle

\begin{abstract}
Global plant maps of plant traits, such as leaf nitrogen or plant height, are essential for understanding ecosystem processes, including the carbon and energy cycles of the Earth system. However, existing trait maps remain limited by the high cost and sparse geographic coverage of field-based measurements. Citizen science initiatives offer a largely untapped resource to overcome these limitations, with over 50 million geotagged plant photographs worldwide capturing valuable visual information on plant morphology and physiology.
In this study, we introduce PlantTraitNet, a multi-modal, multi-task uncertainty-aware deep learning framework that predicts four key plant traits (plant height, leaf area, specific leaf area, and nitrogen content) from citizen science photos using weak supervision. By aggregating individual trait predictions across space, we generate global maps of trait distributions. 
We validate these maps against independent vegetation survey data (sPlotOpen) and benchmark them against leading global trait products. Our results show that PlantTraitNet consistently outperforms existing trait maps across all evaluated traits, demonstrating that citizen science imagery, when integrated with computer vision and geospatial AI, enables not only scalable but also more accurate global trait mapping. This approach offers a powerful new pathway for ecological research and Earth system modeling.
\end{abstract}

\begin{links}
    \link{Code}{github.com/GeoSense-Freiburg/PlantTraitNet}
    \link{Datasets}{huggingface.co/datasets/ayushi3536/PlantTraitNet}
    \link{Extended version}{https://arxiv.org/abs/2511.06943}
\end{links}

\section{Introduction}

Terrestrial plants, as the largest primary producers on Earth, contribute about 60\% to the global net primary productivity~\citep{field1998primary} and play a critical role in the carbon and energy cycles of our Earth system~\citep{pan2011large, schlesinger2020biogeochemistry}. However, understanding how plants influence these cycles is challenging, as the functioning of plants varies profoundly according to their traits. For instance, traits such as canopy height and leaf area control resource acquisition, while leaf tissue properties, such as nitrogen content or dry matter content, are indicators of plant resilience \citep{diaz2016global}. Although these traits are essential for understanding ecosystem processes, the data on such traits is sparse, as their measurement involves costly field surveys and laboratory analysis. Global plant trait databases such as TRY~\citep{kattge2011try} %and BIEN ~\citep{enquist2016cyberinfrastructure} 
aggregate thousands of trait measurements from numerous studies and regions, providing an invaluable resource for functional biogeography and ecosystem modeling. However, even with these collective efforts, significant gaps persist in the geographic coverage of trait data across biomes, ecosystems, and species, constraining our ability to fully understand and predict global patterns of vegetation function and change ~\citep{diaz2016global, kattge2020try}.

Given the strong link between plant morphology and function, plant photographs in concert with computer vision offer a promising avenue for large-scale estimation of plant traits. Citizen science platforms such as iNaturalist~\citep{su2021semi} and Pl@ntNet~\citep{garcin2021plantnet} have collected more than 50 million research-grade plant photographs around the world, creating a unique resource for uncovering global plant trait distributions \citep{wolf2022citizen}.
These datasets, primarily curated for the identification of plant species, provide plant images and species labels, but do not provide trait annotations ~\citep{plantclef-2025, stevens2024bioclip, van2018inaturalist}. However, prior work has demonstrated that trait information for these species can be indirectly obtained by linking species names from citizen science records to trait databases such as TRY~\citep{schiller2021deep, wolf2022citizen}. Through this species-level matching, trait values can be weakly assigned to images, enabling the construction of large-scale, trait-annotated image datasets. These datasets can then be used to train scalable computer vision models for trait prediction from images \citep{schiller2021deep}. 
Such models enable direct trait estimation from photographs, independent of whether the species is known or if a record exists in a trait database. Here, we attempt to advance this approach by predicting multiple traits simultaneously, leveraging shared visual features and underlying trait correlations.
Subsequently, we spatially aggregate trait predictions derived from individual geotagged photographs to create global, gridded geospatial maps representing the trait distributions across plant communities and ecosystems ~\citep{schiller2021deep}.

The geolocation of each photograph not only enables the spatial aggregation of the predictions into geospatial maps but also allows for the integration of spatial context into the prediction process itself ~\citep{schiller2021deep}. For instance, climate, including temperature and precipitation, or phenological information from satellite data is known to be the key variable shaping global trait distributions, making it a promising predictor ~\citep{bruelheide2018global, schiller2021deep, joswig2022climatic}.
However, integrating large-scale geospatial products can be challenging due to data gaps and feature selection \citep{lusk2025smartphones}.

Recent advances in geospatial foundation models (GeoFMs) now support the seamless integration of such context into downstream tasks. Examples include Climplicit \citep{dollinger2025climplicit}, which encodes climate information, and SatCLIP \citep{klemmer2025satclip}, which leverages satellite Earth observation data. Such GeoFMs have demonstrated strong generalization in global mapping applications. In this study, we test the integration of such GeoFMs into visual trait prediction to enhance performance through geospatial context.

An important challenge for computer vision with citizen science data is its inherent heterogeneity and noise \citep{sierra2024divshift}, ranging from inconsistent image quality due to varied photo acquisition methods (feature noise) to ambiguous trait annotations from weak supervision (label noise). Such data characteristics may result in implausible predictions and leave an imprint on the aggregation into global trait products. Moreover, such training data noise can substantially degrade a model's ability to generalize to unseen data ~\citep{lu2022selc, arpit2017closer, zhang2021understanding}.  
To overcome both feature and label noise in the citizen-science data, we propose an uncertainty-aware probabilistic deep learning framework that estimates predictive uncertainty. The predicted uncertainty is used to dynamically down-weight highly noisy samples and to filter out unreliable data points, thereby reducing overfitting to spurious patterns.

Overall, our contributions are summarized as follows:

\begin{itemize}
    \item We introduce the first machine learning–ready dataset that systematically links crowd-sourced plant photographs from citizen science platforms to species-level trait values derived from global trait databases.
    
    \item We present \textbf{PlantTraitNet}, the first uncertainty-aware, multimodal, multi-task deep learning model for global-scale prediction of four key plant traits: height (H), leaf area (LA), specific leaf area (SLA), and leaf nitrogen content (LN).
    
    \item We apply \textbf{PlantTraitNet} on more than 300K independent samples of citizen science photos and spatially aggregate the predictions to global trait maps. A  benchmark against globally distributed vegetation survey data (\textit{sPlotOpen}) revealed that these \textbf{PlantTraitNet}-derived traits maps consistently outperform previous global trait products.

\end{itemize}

\section{Related Work}

Pioneering work by~\citet{schiller2021deep} showed that plant traits, such as height, nitrogen content, specific leaf area, or leaf area, can be predicted from citizen science images using weak supervision, where species-level trait labels are derived from the TRY database~\citep{kattge2020try}.
While ~\citet{schiller2021deep} focused on single-task models, \citet{cherif2023spectra} showed that predicting multiple plant traits simultaneously can exploit trait-trait correlations and shared features in the predictor data.

However, \citet{schiller2021deep} did not assess whether weak supervision enables capturing within-species trait variation (e.g., size differences among individuals of the same species).
Moreover, ~\citet{schiller2021deep} did not test how aggregating individual predictions on a global scale resembles large-scale trait variation across the biosphere. \citet{wolf2022citizen} provided an approach to validate global trait maps using vegetation survey data of plant communities from the collaborative initiative \textit{sPlot} \citep{bruelheide2018global, sabatini2021splotopen} linked with trait data from the TRY database \citep{kattge2020try}. This approach provides an effective means to evaluate the potential of computer vision models for generating trait maps at global scale.

A persistent challenge with citizen science data is the noise in both images and labels~\citep{sierra2024divshift, schiller2021deep}, often structured spatially. Such noise can bias both inference and training, as deep networks tend to memorize noisy labels, compromising generalization~\citep{lu2022selc, arpit2017closer, zhang2021understanding}.

Here, we build on previous work and advance the global trait mapping from citizen science imagery along the following aspects:
\begin{itemize}
\item Using visual and depth-based foundation models~\citep{oquab2023dinov2, yang2024depth} to better represent heterogeneous plant imagery.
\item Leveraging multi-task learning to exploit trait correlations~\citep{cherif2023spectra}.
\item Incorporating uncertainty-aware training to address label noise~\citep{yeo2021robustness, jiang2024uncertainty}.
\item Benchmarking global trait predictions against sPlot vegetation survey data~\citep{wolf2022citizen}.
\item Qualitatively evaluating within-species trait variation.
\item Exploring geospatial fusion to enrich trait mapping.
\end{itemize}

\section{Data}

\begin{figure}[t]
\centering
\includegraphics[width=0.47\textwidth]{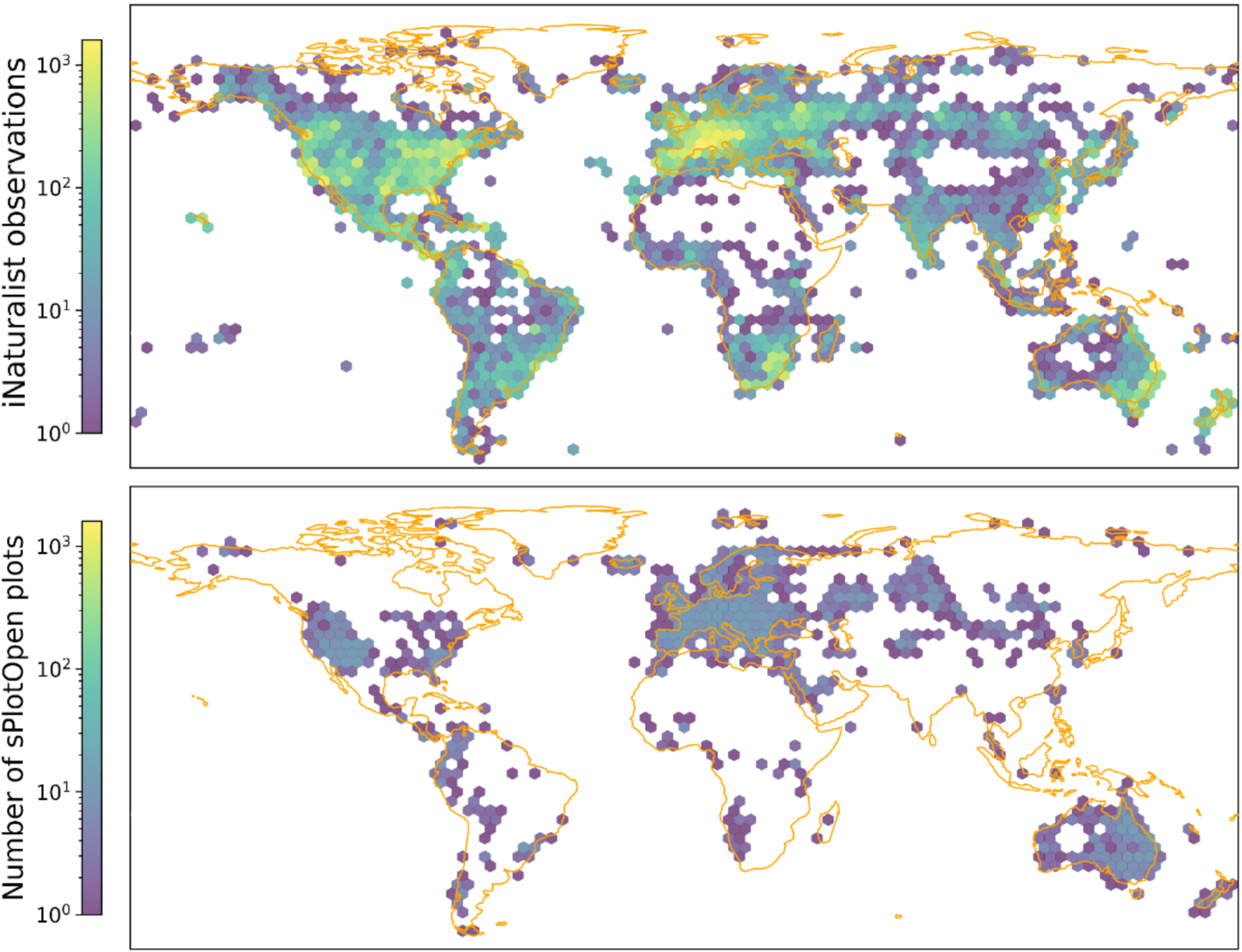}
\caption{Geographic coverage of the citizen science data (top) and independent benchmark reference data (bottom) from vegetation surveys \cite[sPlotOpen,][]{sabatini2021splotopen}.}
\label{fig:spatial}
\end{figure}

\begin{figure}[t]
    \centering
    \includegraphics[width=0.45\textwidth]{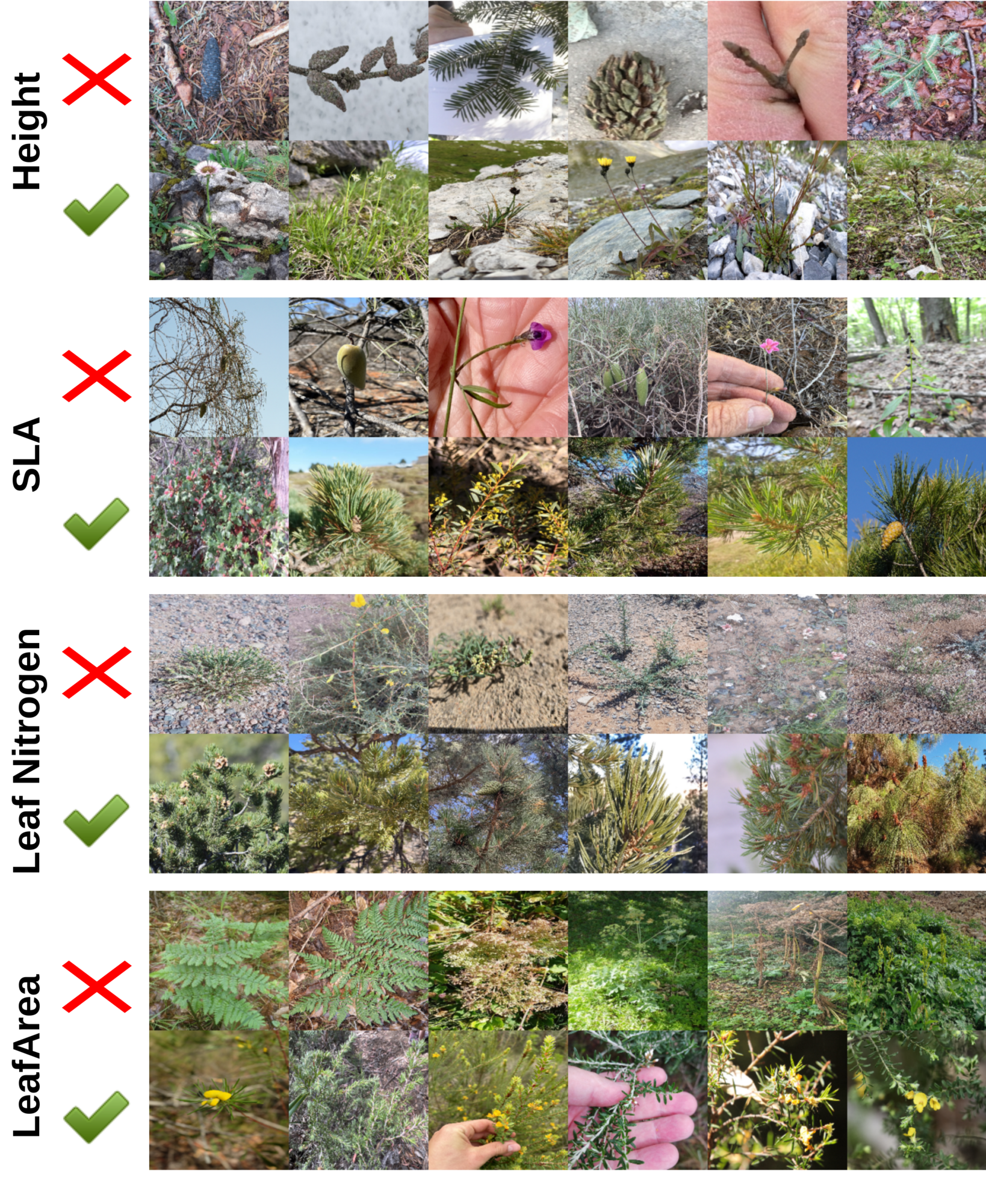} 
   
    \caption{
Randomly sampled images showing highest/lowest predictive uncertainty (see Methodology). \textbf{Observations}: \textbf{Height} uncertainty often from unsuitable contexts (winter scenes, fruits, hands). \textbf{SLA} uncertainty from images lacking visible leaves (bare branches, flowers, buds). \textbf{Leaf Nitrogen}: low-quality/blurry images. \textbf{Leaf Area}: exotic leaf types (e.g., ferns).
}
    \label{fig:uncer_cerimg}
\end{figure}

\subsection{Weakly Labeled Citizen Science Photographs}
To predict plant traits at a global scale, we utilize two large-scale citizen science datasets: iNaturalist~\citep{gbif2025, su2021semi} and Pl@ntNet-300K~\citep{garcin2021plantnet}. These datasets consist of plant images annotated with species labels and geolocations but lack direct trait measurements. 
Following~\citet{schiller2021deep}, we weakly annotate each image using species-level trait distributions from the TRY database~\citep{kattge2020try}, based on the premise that interspecific trait variation (variation between species) generally exceeds intraspecific variation (variation within species)~\citep{dong2020components, wright2017global}.

We model each trait as a normal distribution per species, using TRY-derived means and standard deviations, and sample trait values within the interquartile range to reduce outlier influence. To account for intraspecific variability, we resample traits for each image at every training epoch ~\cite{schiller2021deep}.

This weak supervision introduces label noise, especially for traits with strong intraspecies variability across developmental stages (e.g., juvenile trees assigned mature height). 

We further reduce noise through model-driven uncertainty estimates (see Methodology). The final dataset includes approximately 220K training images across 5K species and over 80K images in the validation set.

\subsection{Vegetation Survey Data - sPlotOpen}
For evaluation, we use the sPlotOpen database \cite{sabatini2021splotopen}. The georeferenced sPlotOpen records represent plant community compositions, which were linked with trait data from the TRY database \citep{kattge2020try}.  This data provides global trait maps of community-weighted mean (CWM) trait values.

\subsection{Reference Data}
To aid uncertainty-based filtering, we curated a small dataset of ~780 species with images and trait measurements taken from the same individual at the same time including observations from diverse regions such as Germany, La Palma, India, Australia etc. (See Appendix for details)

\section{Methodology}
\begin{figure}[th] % 'h' means place the figure approximately here
    \centering
    \includegraphics[width=0.47\textwidth]{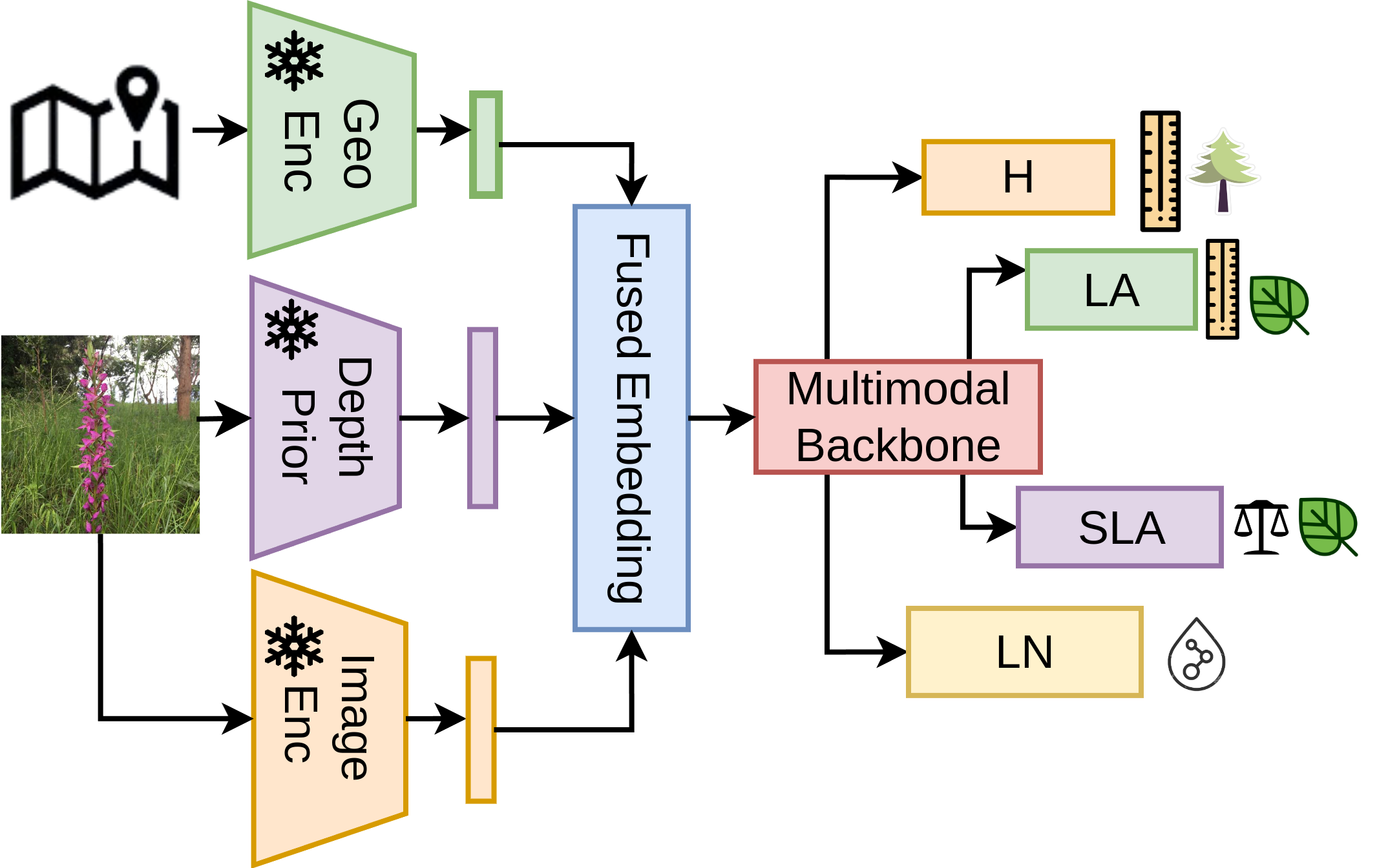} 
    \caption{The model integrates image, depth, and geospatial embeddings. These are fused within a multimodal backbone, which then uses individual heads to predict height (H), leaf area (LA), specific leaf area (SLA), and leaf nitrogen (LN).}

    \label{fig:arch_plantraitnet}
\end{figure}
\begin{figure*}[tp] %  
\centering
\includegraphics[width=\linewidth]{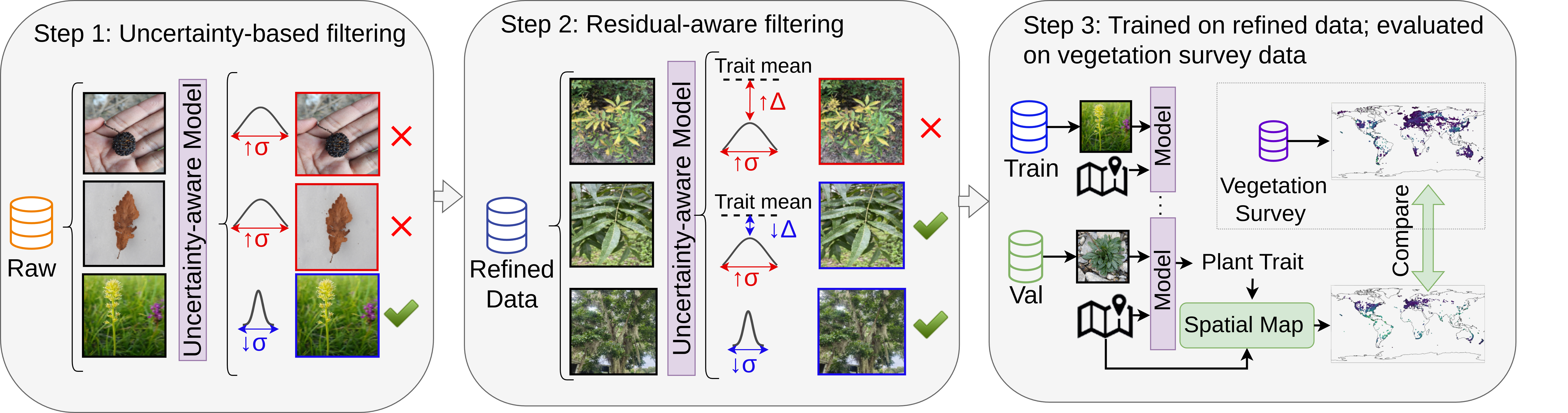} 

\caption{Overview of the pipeline. We filter weakly labeled citizen science data (Raw data) based on high model uncertainty (Step 1) and large residuals from species trait medians (Step 2). We use this refined data for training the models (Step 3), which are evaluated by comparing spatially aggregated predictions (1° resolution) against overlapping vegetation surveys (sPlotOpen).}
\label{fig:datacleaningloop}
\end{figure*}
The PlantTraitNet architecture (Fig. \ref{fig:arch_plantraitnet}), uses a general-purpose vision encoder. In addition to image features, we incorporate depth and geospatial priors. These modality-specific embeddings are fused using simple concatenation. The fused representation is passed through a shared multimodal backbone, followed by trait-specific linear heads.

\subsection*{Image Encoder}

Given an input image \(\mathbf{I} \in \mathbb{R}^{3 \times H \times W}\), we use the pretrained DINOv2 ViT-B/14 encoder to extract a sequence of patch-level feature embeddings $ \in \mathbb{R}^{N \times C}$  ~\citep{oquab2023dinov2}. 

We apply adaptive average pooling along the patch tokens $N$, reducing it to a 32-dimensional representation. This pooling operation is parameter-free and preserves condensed spatial structure before projection.
This pooled output is flattened and passed through a multi-layer perceptron (MLP) to generate the embedding of dimension 768.

\subsection{Depth Priors from Foundation Models}

A novel addition to our architecture is the use of depth priors from foundation models for monocular depth estimation. While standard 2D RGB images lack explicit three-dimensional spatial cues, depth information encodes the distance between the sensor and surface points on the plant, enabling a more accurate reconstruction of plant morphology and structure.
To incorporate depth, we use the pretrained and frozen encoder from the Depth-Anything-V2 (DA-V2) model ~\cite{yang2024depth}, denoted as $h$.
%We encode the RGB image $I$.
Although various models could be used, we adopt DA-V2 for its strong generalization capabilities, attributed to its training on large-scale labeled and unlabeled datasets and its student-teacher distillation framework. We use the ViT-B variant, which outputs a set of embeddings $h(I) \in \mathbb{R}^{N \times C}$.
% , where $D = 768$
Similar to the image encoder, we apply adaptive average pooling along the patch token dimension $N$, reducing it to a 64-dimensional representation. This pooled output is flattened and passed through a MLP to generate the depth prior embedding of dimension 768.

\subsection{Geospatial Priors from Foundation Models}

Plants are tailored to local climatic conditions, such as precipitation and temperature, through their traits \citep{joswig2022climatic}.

To incorporate this climatic context as a cue in the prediction process, we integrate Climplicit \citep{dollinger2025climplicit} into our architecture, a spatio-temporal geo-location encoder trained on the CHELSA climate dataset \citep{karger2017climatologies}. Climplicit maps latitude, longitude, and month of the year to a continuous embedding that implicitly captures climatic factors such as temperature and precipitation.
%This allows the model to encode fine-grained climatic variation, even in the absence of explicit climate variables.
To incorporate seasonal trends, we concatenate the embeddings for the months of March, June, September, and December.

\subsection{Multimodal and Multi-Task Backbone}
% \subsection{Multimodal Fusion Architecture}
Let \(\mathbf{X}_{\text{img}} \in \mathbb{R}^{768}\) denote the image embedding obtained from the pretrained DINOv2 encoder, and \(\mathbf{X}_{\text{depth}} \in \mathbb{R}^{768}\) denote the depth embedding obtained from the DA-V2 encoder. To incorporate geospatial context, we project the 1024-dimensional embedding produced by Climplicit denoted as \(\mathbf{X}_{\text{geo}} \in \mathbb{R}^{1024}\) to a 256-dimensional vector using a trainable linear projection.

The multimodal representation is formed by concatenating all embeddings and is then projected to a 1024-dimensional representation via a linear layer: $
\mathbf{Z}=\text{Proj}( \text{concat}(\mathbf{X}_{\text{img}}, \mathbf{X}_{\text{depth}}, \text{Proj}(\mathbf{X}_{\text{geo}}))) \in \mathbb{R}^{1024}.
$
The resulting embedding is passed through a residual network of 8 residual blocks with hidden dimension twice the embedding size.
This architecture and embedding dimensions were chosen based on an ablation across multiple configurations (see Appendix). Finally, the output feature representation is passed to four independent heads for trait prediction in our multi-task architecture.

\subsection{Uncertainty Estimation}

To capture uncertainty in plant trait prediction, each trait-specific prediction head outputs both the predicted value and its associated uncertainty, following the method by \citep{jiang2024uncertainty}. For each trait \( m \in \{1, \ldots, M\} \), the model predicts two values for each sample \( n \): the mean \( \hat{\mu}_n^m \) and the log-scale parameter \( \hat{s}_n^m \), where the scale or standard deviation is given by \( b_n^m = \exp(\hat{s}_n^m) \).

We model the predictive distribution differently for each trait based on its statistical characteristics. For Leaf Area (LA), which exhibits a long-tailed distribution, we use a Laplace distribution parameterized by mean \( \hat{\mu}_n^m \) and scale \( b_n^m = \exp(\hat{s}_n^m) \). The Laplace distribution is more suitable for modeling long-tailed distributions compared to Gaussian distributions ~\cite{jiang2024uncertainty}.

For the remaining traits, Height (H), Specific Leaf Area (SLA), and Leaf Nitrogen (LN), we assume a Gaussian distribution with mean \( \hat{\mu}_n^m \) and standard deviation \( \sigma_n^m = \exp(\hat{s}_n^m) \). Although plant height has strong skewness (dominance of small plants), we employ stratified sampling based on plant functional types during training. This ensures that each mini-batch contains approximately equal representation of grasses, shrubs, and trees, which may make the Gaussian assumption more suitable for modeling this trait (respective ablations are described in the Appendix).

\section{Uncertainty-Guided Data Cleaning Loop}

\label{sec:uncerguideddatacleaning}

Citizen science image datasets offer large-scale and diverse data for plant trait modeling but suffer from substantial noise and inconsistencies. Common issues include the presence of non-plant objects, non-representative plant parts, uninformative specimens, and scenes that are too dense, distant, or blurred (Fig.~\ref{fig:uncer_cerimg}). Manual curation at this scale is impractical, and species-level trait annotations often ignore individual variation. Neural networks often memorize noisy labels which harms generalization and makes noise handling essential~\citep{lu2022selc, arpit2017closer, zhang2021understanding}. ~\citet{lu2022selc} shows models initially learn from clean samples but, past a `turning point', begin memorizing noise, leading to poorer generalization.

Building on this insight, we implement a two-step data cleaning loop guided by model-predicted uncertainty (Fig.~\ref{fig:datacleaningloop}). The first step applies uncertainty-based filtering: after early training for a single epoch on raw data, all training images are inferred and ranked by trait-wise uncertainty, and those exceeding a joint threshold across all traits are filtered out. This process continues iteratively until the number of samples jointly flagged as uncertain across traits falls below a predefined threshold.

However, uncertainty alone can be unreliable for heteroscedastic traits such as plant height, where variance naturally increases with trait magnitude. In such cases, high uncertainty may reflect genuine biological variability rather than label noise. Consequently, filtering solely by uncertainty risks biasing the cleaned dataset toward lower-variance samples. To mitigate this, the second stage performs residual-aware filtering, combining uncertainty and prediction residuals. For this, we identify the `turning point' while training for each trait. We do so by tracking performance on the reference dataset and selecting the epoch after which trait-wise performance begins to deteriorate. Using predictions from this epoch, we calculate the mean absolute error between predicted trait values and species-level means for samples with high uncertainty. Images with high uncertainty and large residuals are filtered from the dataset. The cleaning loop terminates when the number of samples satisfying the filtering criteria becomes negligible. Further details are provided in Appendix. 

\section{Model Evaluation and Selection}

Following the approach of previous studies \citep{wolf2022citizen, dechant2024intercomparison}, we aggregated the plot-level trait values from sPlotOpen to a 1-degree spatial resolution to generate a global benchmark dataset. We then applied PlantTraitNet to predict trait values using more than 300K globally distributed citizen science observations. The predicted trait values were aggregated to the same 1-degree resolution and filtered to include only grid cells with at least 20 observations, resulting in over 890 grid cells. To ensure a robust and interpretable evaluation, we used complementary metrics that capture different aspects of model performance: the coefficient of determination (R²) quantifies the proportion of variance in observed trait values explained by the predictions, the normalized mean absolute error (nMAE) measures the average deviation normalized by the observed trait range, and Pearson’s correlation coefficient (r) on log-transformed values quantifies the linear relationship between predicted and observed data.
To account for spatial sampling bias, all metrics are weighted by the area of each 1-degree grid cell. We also compare previously published trait maps against the same sPlotOpen CWM values on overlapping grid cells \cite{boonman2020assessing, butler2017mapping, madani2018future, moreno2018methodology, schiller2021deep, van2014fully, wolf2022citizen}. For the ablation study, all models were assessed solely on our validation dataset.

The final model, with approximately 90M trainable parameters, was trained for up to 30 epochs with a batch size of 256 on a single NVIDIA RTX A6000 GPU (using approximately 20~GB VRAM). 
To select the optimal model checkpoint across all traits, we compute the Pareto front using the Non-Dominated Sorting (NDS) algorithm~\citep{Deb2002NSGAII}. 
We then calculate the hypervolume for all candidate checkpoints on this front~\citep{Zitzler2001SPEA2} and select the checkpoint that maximizes the hypervolume. 
Following~\citet{lacoste2019quantifying}, we estimate a total of 93.86~kg~CO$_2$ emissions for all experiments across seeds. 
This estimate excludes testing and failed runs and therefore likely underestimates the total emissions, but it provides a reasonable guideline for future model training.

\section{Results}
\begin{table}[ht]
\centering
\begin{tabular}{l l c c c c}
\toprule
Method & Metric & H & LA & SLA & LN \\
\midrule
\multirow{1}{*}{Ours} 
& $R^2$ ↑ & \textbf{0.19} & \textit{0.30} & \textit{0.23}& -0.16
\\
(Raw)&
nMAE ↓& \textbf{0.22} & \textbf{0.14 } & \textit{0.14} & \textit{0.17} \\ 
& $r$ ↑& \textbf{0.45} & \textit{0.56} & \textbf{0.59} & \textit{0.49} \\ 
\midrule
\multirow{1}{*}{Ours} 
& $R^2$ ↑   & \textit{0.18} & \textbf{0.34} & \textbf{0.27} & \textit{-0.12 } \\
(Refined)& nMAE ↓    & \textbf{0.22} & \textbf{0.14} & \textbf{0.13} & \textit{0.17} \\ 
& $r$ ↑     & \textbf{0.45} & \textbf{0.57 } & \textbf{0.59 } & \textbf{0.50 } \\ 
\midrule
\multirow{3}{*}{Schiller}
& $R^2$ ↑ & -0.32  & 0.11 & 0.16 & \textbf{0.06} \\
& nMAE ↓ & \textit{0.28} & \textit{0.17} & \textit{0.14}  & \textbf{0.14} \\
& $r$ ↑ & 0.42 & 0.52 & \textit{0.53}& 0.40 \\
\midrule
\multirow{3}{*}{Wolf}
& $R^2$ ↑ & -0.61 & -0.02 & 0.02 & -0.20 \\
& nMAE ↓ & 0.31  & 0.18 & 0.16 & 0.18  \\
& $r$ ↑ & \textit{0.43} & 0.53  & 0.50 & 0.41 \\
\midrule
\multirow{3}{*}{Moreno}
& $R^2$ ↑ & -- & -- & -0.72  & -0.85 \\
& nMAE ↓ & -- & -- & 0.23  & 0.22 \\
& $r$ ↑ & -- & -- & 0.23  & 0.17  \\
\midrule

\multirow{3}{*}{Butler}
& $R^2$ ↑ & -- & -- & -0.17  & -0.50 \\
& nMAE ↓ & -- & -- & 0.18 & 0.20 \\
& $r$ ↑ & -- & -- & 0.29 & 0.32\\
\midrule
\multirow{3}{*}{Boonman}
& $R^2$ ↑ & -- & -- & 0.03 & -0.37  \\
& nMAE ↓ & -- & -- & 0.16 & 0.18 \\
& $r$ ↑ & -- & -- & 0.49 & 0.20 \\
\midrule
\multirow{3}{*}{Madani}
& $R^2$ ↑ & -- & -- & -0.76 & -- \\
& nMAE ↓ & -- & -- & 0.23 & -- \\
& $r$ ↑ & -- & -- & -0.07 & -- \\
\midrule
\multirow{3}{*}{Bodegom}
& $R^2$ ↑ & -- & -- & -1.00  & -- \\
& nMAE ↓ & -- & -- & 0.24 & -- \\
& $r$ ↑ & -- & -- & 0.33 & -- \\

\bottomrule
\end{tabular}
\caption{
Global trait map benchmarking against sPlotOpen CWMs (1\textdegree{} resolution). \textbf{Best}. \textit{Second-best}. External products: Schiller~\cite{schiller2021deep}, Wolf~\cite{wolf2022citizen}, Moreno~\cite{moreno2018methodology}, Butler~\cite{butler2017mapping}, Boonman~\cite{boonman2020assessing}, Madani~\cite{madani2018future}, Bodegom~\cite{van2014fully}.
PlantTraitNet (Ours) is evaluated after training on both raw and refined datasets.}
\label{tab:globalbenchmark}

\end{table}

% \begin{figure}[htp]
% \centering
% \includegraphics[width=0.47\textwidth]{taxonomic_tree.pdf}
% \caption{Mean relative prediction error (MRPE) computed on validation data at the family level, visualized along the taxonomic tree, for height (H), leaf area (LA), specific leaf area (SLA) and leaf nitrogen (LN).}
% \label{fig:phylo}
% \end{figure}

% \begin{figure*}[htp]
% \centering
% \includegraphics[width=0.71\textwidth]{taxonomic_tree.pdf}
% \caption{Mean relative prediction error (MRPE) computed on validation data at the family level, visualized along the taxonomic tree, for height (H), leaf area (LA), specific leaf area (SLA) and leaf nitrogen (LN).}
% \label{fig:phylo}
% \end{figure*}

\begin{figure*}[!tb]
\centering
\includegraphics[width=0.75\textwidth]{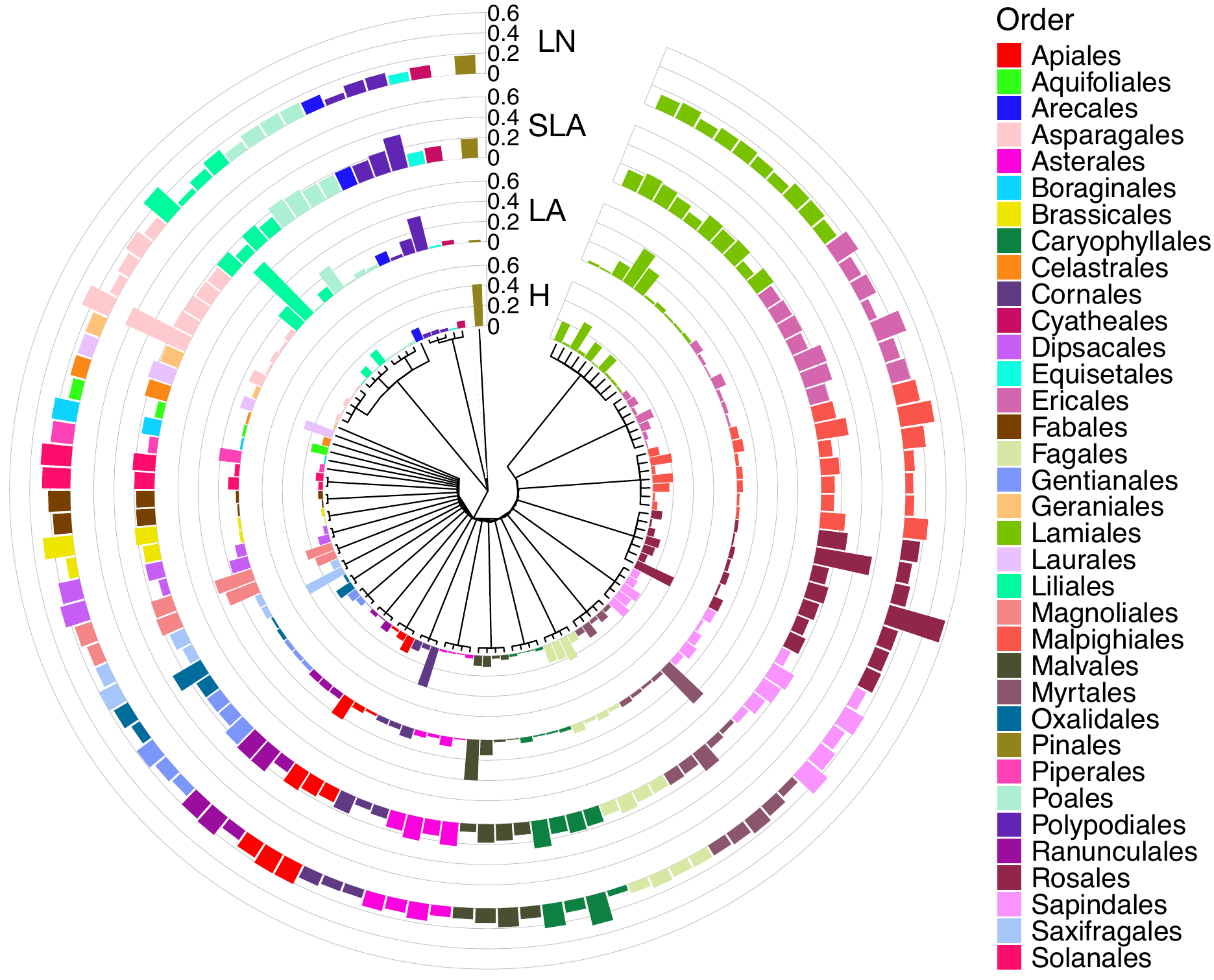}
\caption{Mean relative prediction error (MRPE) computed on validation data at the family level, visualized along the taxonomic tree, for height (H), leaf area (LA), specific leaf area (SLA) and leaf nitrogen (LN).}
\label{fig:phylo}
\end{figure*}

\begin{table*}[!htb]
\centering
\begin{tabular}{c| c| c| c| c c c c | c}
\hline
& Image & Geo & Depth & H & LA & SLA & LN & \# Top ranks \\
\hline
\multirow{8}{*}{\rotatebox{90}{Multi-Task}}  
&DinoV2 & \nomark & \nomark 
& 0.15 ± 0.00 & 0.31 ± 0.00 & \textbf{0.32 ± 0.00} & 0.14 ± 0.01 & 1 \\
\cline{2-9}
&BioCLIP & \nomark & \nomark 
& 0.15 ± 0.01 & 0.3 ± 0.00 & \textbf{0.32 ± 0.01} & 0.15 ± 0.04 & 1 \\
\cline{2-9}
&DinoV2 & SatCLIP & \nomark 
& 0.16 ± 0.02 & 0.27 ± 0.04 & 0.25 ± 0.02 & 0.11 ± 0.05 & 0 \\
\cline{2-9}
&DinoV2 & GeoCLIP & \nomark 
& \textit{0.17 ± 0.01} & \textbf{0.33 ± 0.01} & \textbf{0.32 ± 0.00} & 0.15 ± 0.02 & 3 \\
\cline{2-9}
&DinoV2 & Climplicit & \nomark 
& \textbf{0.19 ± 0.01} & \textit{0.32 ± 0.01} & \textit{0.31 ± 0.01} & 0.16 ± 0.06 & 3 \\
\cline{2-9}
&BioCLIP & Climplicit & \nomark 
& \textbf{0.19 ± 0.00} & \textit{0.32 ± 0.02} & \textit{0.31 ± 0.01} & 0.15 ± 0.06 & 3 \\
\cline{2-9}
&BioCLIP & Climplicit & DA-V2 
& 0.16 ± 0.01 & 0.28 ± 0.03 & 0.30 ± 0.00 & \textbf{0.19 ± 0.02} & 1 \\
\cline{2-9}
&DinoV2 & Climplicit & DA-V2 
& \textbf{0.19 ± 0.02} & \textit{0.32 ± 0.01} & \textit{0.31 ± 0.02} & \textit{0.18 ± 0.05} & 4 \\
\hline
\rotatebox{90}{ST} & DinoV2 & Climplicit & DA-V2 
& 0.12 ± 0.01 & \textbf{0.34 ± 0.01} & \textbf{0.33 ± 0.01} & \textbf{0.21 ± 0.02} & -- \\
\hline
\end{tabular}
\caption{
Multi-modal ablation study for plant trait prediction. Results are reported as mean $R^2 \pm$ 1 standard deviation over 3 runs.
\textbf{Bold} indicates the best result, and \textit{italic} indicates the second-best. `\# Top ranks' counts the number of top-two rankings. The last row reports the performance of the best multi-task model when evaluated in a single-task (ST) setting.
}
\label{tab:ablation_summary}
\end{table*}

\subsection{Benchmarking Against Vegetation Survey Data}
We benchmark our global trait maps derived from models trained on both raw and filtered data and those of previous studies using R², Pearson’s $r$, and nMAE (Table~\ref{tab:globalbenchmark}). Depending on the metric, our model for LN delivers comparable results with those by Schiller et al.~\cite{schiller2021deep}. For the other three traits, our models consistently achieve higher performance than previously published plant trait maps.  The overall improvement reflects the model’s ability to capture complex and variable patterns in large-scale trait prediction. While the $r$ scores suggest that the maps capture relative differences, substantially lower R² scores indicate that PlantTraitNet maps and all other products are systematically biased (also see Appendix), underscoring the inherent challenge of revealing morphological and physiological ecosystem patterns at global scale.

\begin{figure*}[!ht]
\centering
\includegraphics[width=\textwidth]{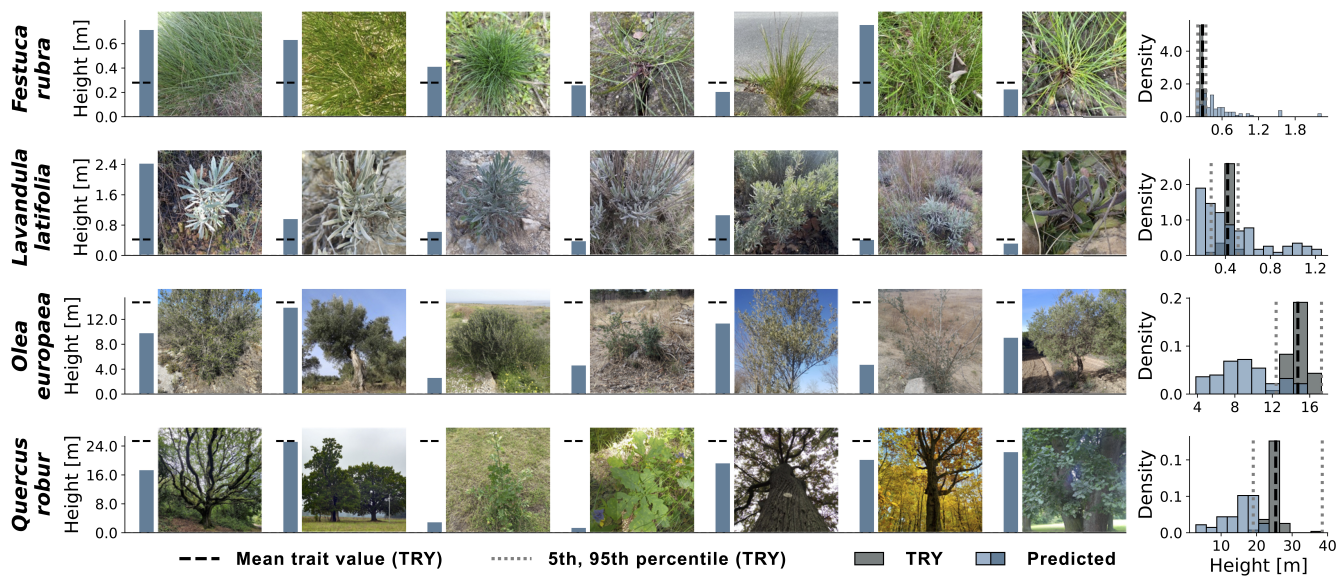}
\caption{Intraspecific variation in predicted height for four species. Bar plots (left) show model predictions; histograms (right) show height distributions predicted from up to 100 images compared with up to 100 measurements from the TRY database.}
\label{fig:quali-height}
\end{figure*}

\subsection{Model Performance Across and Within Species}

To reveal trait variability within the biosphere on a global scale, a computer vision model must be robust across species, and thus phylogenetic lineages. Using  inferences and species information from the validation data, we show that the residuals of PlantTraitNet are largely unsystematically distributed in the phylogenetic space (Fig.~\ref{fig:phylo}). To quantify this, we use two standard metrics of phylogenetic signal:

Pagel’s $\lambda$, which captures broad-scale phylogenetic autocorrelation in residual covariance \citep{pagel1999inferring} and Blomberg’s $K$, which is more sensitive to fine-scale signals among closely related species \citep{blomberg2003testing}. For SLA and leaf nitrogen, the phylogenetic signal is weak ($\lambda$ = 0.04 and 0.15; $K$ = 0.0053 and 0.0076), suggesting that prediction errors are largely independent of species relatedness. Although errors for height ($\lambda$ = 0.80, $K$ = 0.018) and leaf area ($\lambda$ = 0.56, $K$ = 0.0067) show some phylogenetic autocorrelation, the consistently low $K$ values indicate that even closely related species do not share systematic prediction biases (see Appendix for details). Although PlantTraitNet was trained using weak annotations at the species level, these findings underscore the model’s strong generalizability and robustness across the plant tree of life.

Despite weak supervision at the species level, PlantTraitNet captures within-species variability in trait expression. This is particularly evident in the case of height prediction (Fig.~\ref{fig:quali-height}), where the model reflects differences across growth forms and developmental stages within individual species (see Appendix for additional examples of within-species variability across traits). This suggests that the model is not simply regressing to a species-level mean but is sensitive to morphological cues in the images that reflect ecological and ontogenetic variation.

\subsection{Effect of Input Modalities}

To evaluate the contribution of each input modality to trait prediction, we conduct an ablation study using different combinations of image, geospatial, and depth information (Table~\ref{tab:ablation_summary}). Our goal is to understand how each modality influences model performance across key plant functional traits: H, LA, SLA, and LN. For this ablation study, we also experimented with a pretrained BioCLIP \citep{stevens2024bioclip} encoder as an alternative to DinoV2. For BioCLIP, we extracted the embedding from its classification token, as it empirically showed superior performance (detailed in the Appendix). For geospatial priors, we assess SatCLIP~\citep{klemmer2025satclip}, which is trained on satellite imagery and captures vegetation density and phenology; GeoCLIP~\citep{vivanco2023geoclip}, a geo-localization model trained on natural images; and Climplicit ~\citep{dollinger2025climplicit}

We find that image features alone provide a strong baseline, with DINOv2 and BioCLIP performing comparably. Adding geospatial priors from Climplicit consistently improves performance across traits, reflected in higher R². Adding depth information on top of the image and climate input leads to marginal changes overall. In general, image features provide a strong foundation for trait inference, while the integration of climate information significantly enhances prediction. Although depth contributes selectively, its inclusion offers a modest gain in average performance, supporting the use of all three modalities in the final model.

\subsection{Multi-Task versus Single-Task}
In Table~\ref{tab:ablation_summary}, we also compare the effect of jointly predicting all traits (multi-task) versus independently predicting each trait using the same architecture with single trait heads (single-task). While the single-task architecture yields marginally better performance for LA, SLA, and LN (e.g., higher $R^2$ and lower nMAE), the multi-task model shows a substantial performance gain for H, improving $R^2$ from 0.12 to 0.19. Importantly, the multi-task model achieves these results with significantly lower computational cost—training a single joint model instead of four separate ones reduces training time and GPU memory usage by approximately 75\%. Thus, the multi-task model provides a better overall balance of performance and efficiency.

\section{Discussion}

Predicting plant traits from citizen science photos is challenging due to data variability and biases, including spatial and taxonomic bias, and overrepresentation of smaller growth forms such as grasses and herbs~\citep{di2021observing, sierra2024divshift}. Ecological complexity adds difficulty, as traits vary across biomes, with generalists showing common traits and specialists distinct ones, resulting in skewed, long-tailed distributions (see Appendix). Unlike animals with fixed body plans, e.g. with symmetric and fixed numbers of legs or arms, plants have a comparably flexible morphology, resulting in varying numbers of plant organs, such as leaves or branches. This variability complicates trait prediction via computer vision. Despite these challenges, our results show promising potential. Future work should focus on reducing biases through targeted data acquisition. Increased acquisition of reference data to enable better ‘turning point’ selection and incorporate label correction to improve model robustness and generalization in ecological contexts.

\section{Conclusion}

Our understanding of plant–environment interactions is limited by the sparse geographic and taxonomic coverage of morphological and physiological trait data. We demonstrate that citizen science plant images, combined with machine learning can be used to predict and map global distributions of key ecological plant traits using only geolocated images making the approach highly scalable across biomes.
Despite relying on weak supervision via species-level trait annotations, our models capture consistent intraspecific variation. Integrating geospatial context through Earth observation foundation models (GeoFMs) and structural cues via depth priors improves predictive performance and model robustness.
Our multi-task framework enables simultaneous prediction of multiple traits, capturing inter-trait dependencies while improving computational efficiency. Benchmarking against existing global trait maps shows that our approach achieves state-of-the-art performance. This establishes a new baseline for large-scale trait inference from image data, offering a powerful alternative to traditional mapping based on field sampling and extrapolation.
By leveraging abundant publicly available plant images, our method enables automated, global retrieval of core traits, offering new opportunities to explore functional diversity and improve ecosystem modeling under global change.

\section*{Acknowledgements}
This study was funded by the German Research Foundation (DFG) within the project PANOPS (Revealing Earth’s plant functional diversity with citizen science; project no. 504978936)

\bibliography{aaai2026}
\newpage
\input{sec/supplementarymaterial}

\end{document}

%% file: sec/supplementarymaterial.tex
\section{Supplemental Material}
\subsection{Data}
\subsubsection{Citizen Science Data Preparation}
\label{appendix:dataprocessing}
The preparation of our citizen science training data involved several preprocessing steps to ensure quality and ecological representativeness. Using the geolocation metadata of the images, we removed samples from aquatic and urban areas based on the ESA WorldCover 10m v100 land cover map \cite{zanaga2021esa}. 

We weakly annotate these images with species-level trait values from the TRY database \citep{kattge2020try}. To reduce label noise due to sparsely represented species, we excluded species with fewer than three trait observations in the TRY database. Additionally, to further mitigate the impact of outliers, we computed the species-level median trait values after removing observations below the 5th and above the 99th percentiles.

Finally, we constructed the training and validation splits using an 80-20 partition, stratified by plant growth form (trees, shrubs, grasses) to preserve diversity across the subsets. Information on plant growth form was derived from the TRY database \citep[][Trait-ID 4]{kattge2020try}. For each species, we standardized the provided  growth forms into 3 classes (Tree, Shrub, Grassland) and subsequently applied a majority vote per species \cite{lusk2025smartphones}.

For benchmarking PlantTraitNet against other products, we jointly evaluate the model on a comprehensive evaluation set. This benchmark consists of a large-scale, uncurated set of 300K randomly sampled citizen science images, which were downloaded without any filtering for species name or quality, and about 80K images from the validation split of our curated training data.
% Fig. \ref{fig:traitdistribution} shows distribution of each trait.

\begin{figure*}[ht]
\centering
\includegraphics[width=\textwidth]{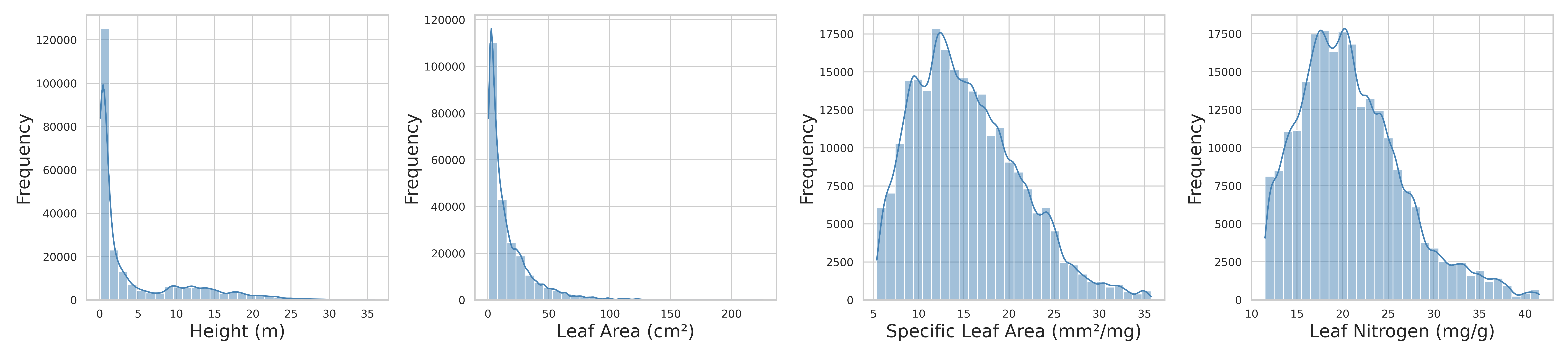}
\caption{Distribution of median trait values at the species level in the weakly labeled citizen science training data.}
\label{fig:traitdistribution}
\end{figure*}

\subsubsection{Reference Data to Aid Uncertainty-Guided Data Cleaning}

\begin{figure}[ht]
    \centering
    \includegraphics[width=\linewidth]{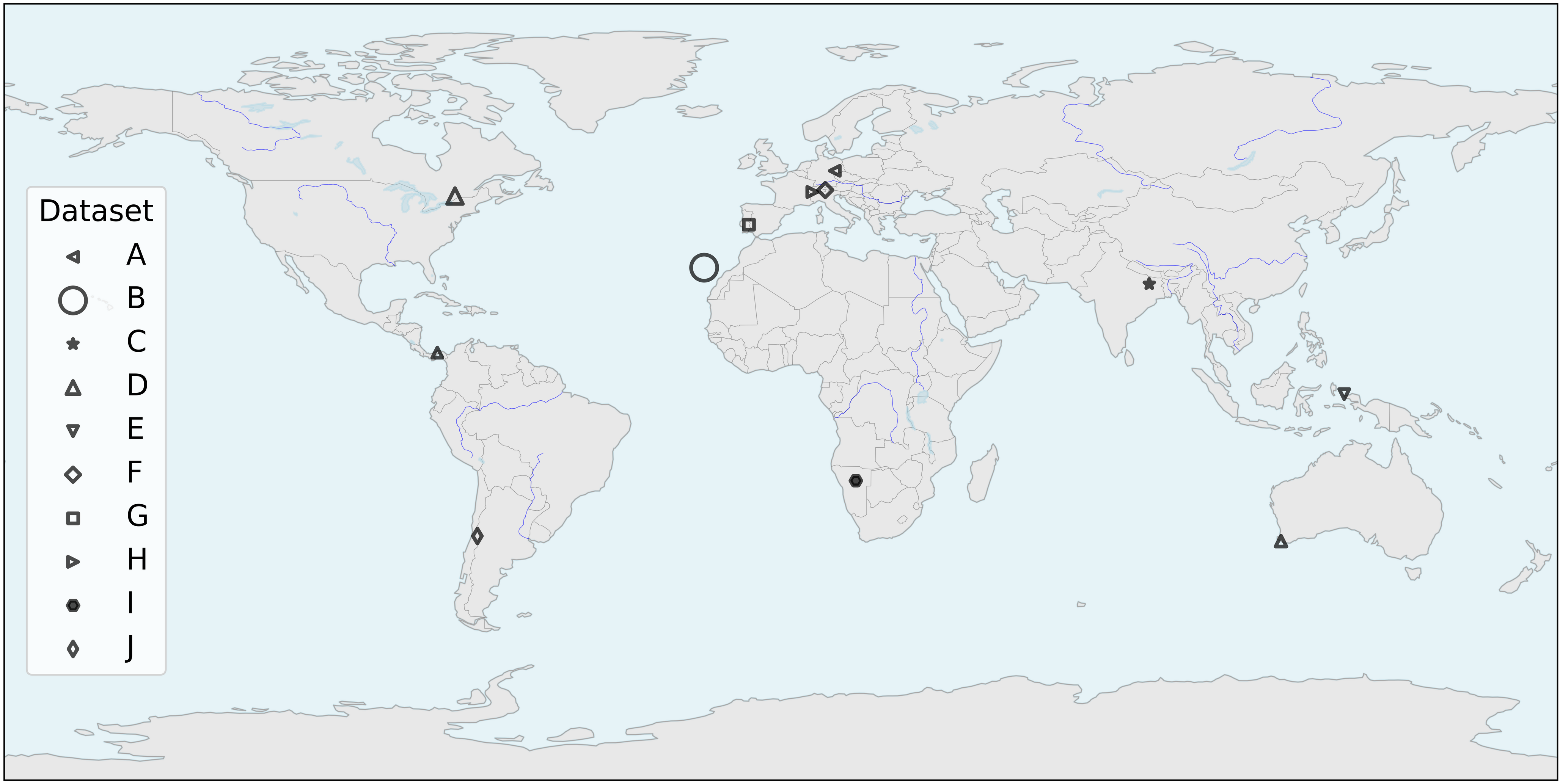}
    \caption{Locations of scientifically curated reference datasets. Marker symbols indicate distinct test data sources from Germany (A), La Palma (B), India (C), Australia (D), Panama (D), Canada (D), Indonesia (E), Switzerland (F), Portugal (G), Switzerland (H), Namibia (I), and Chile (J). Marker sizes are proportional to the corresponding dataset size.}
    \label{fig:real-test-loc}
\end{figure}

Our uncertainty-guided data cleaning strategy is grounded in the concept of a turning point—a stage during training when the model transitions from learning generalizable patterns to overfitting noisy labels \cite{lu2022selc}. Detecting this turning point is critical for effective uncertainty-aware data cleaning. However, in weakly supervised settings, where ground truth labels are absent or imprecise, identifying this transition reliably is challenging. To overcome this limitation, we curated a high-quality reference dataset consisting of plant images paired with accurate trait measurements.
%Motivated by the goal of identifying a `turning point' to support uncertainty-aware data cleaning, we curated a reference dataset.
This dataset consists of smaller datasets distributed globally from various previous studies and a range of collaborators. Each collaborator contributed plant trait measurements and associated images from ecologically diverse locations, including Germany, La Palma, India, Australia, Panama, Canada, Indonesia, Switzerland, Portugal, and Namibia (see \Cref{fig:real-test-loc}). Detailed statistics on sample sizes and taxonomic diversity, quantified as the number of unique species, are reported in Table~\ref{tab:testtraitsmeta}. Because different collaborators focused on specific traits, we maintain separate datasets per trait.

\begin{table}[ht]
    \centering
    \scriptsize
    \caption{Metadata of the curated plant reference dataset to aid uncertainty-guided data cleaning.}
    \label{tab:testtraitsmeta}
    \renewcommand{\arraystretch}{1.25}
    \begin{tabular}{l|l|c|c}
        \hline
        \textbf{Trait} & \textbf{Units} & \textbf{\#Data points} & \textbf{\#Species} \\ 
        \hline
        Plant height (H) & m & 4478 & 452 \\ 
        \hline
        Leaf Area (LA) & cm\textsuperscript{2} & 2386 & 136 \\ 
        \hline
        Specific Leaf Area (SLA) & mm\textsuperscript{2}/mg & 1796 & 199 \\ 
        \hline
        Leaf Nitrogen Content (LN) & mg/g & 773 & 73 \\
        \hline
    \end{tabular}
\end{table}

\subsection{Methodology}
\subsubsection{Uncertainty Estimation}

To capture predictive uncertainty across traits, we model the output distribution for each trait type, denoted as \( m \in \{1, \ldots, M\} \). Due to its long-tailed nature, Leaf Area (LA) is modeled using a Laplace distribution, parameterized by a predicted mean \( \hat{\mu}_n^m \) and scale \( b_n^m = \exp(\hat{s}_n^m) \), where \( \hat{s}_n^m \) is the predicted log-scale.

\begin{figure}[ht]
\centering
\includegraphics[width=\linewidth]{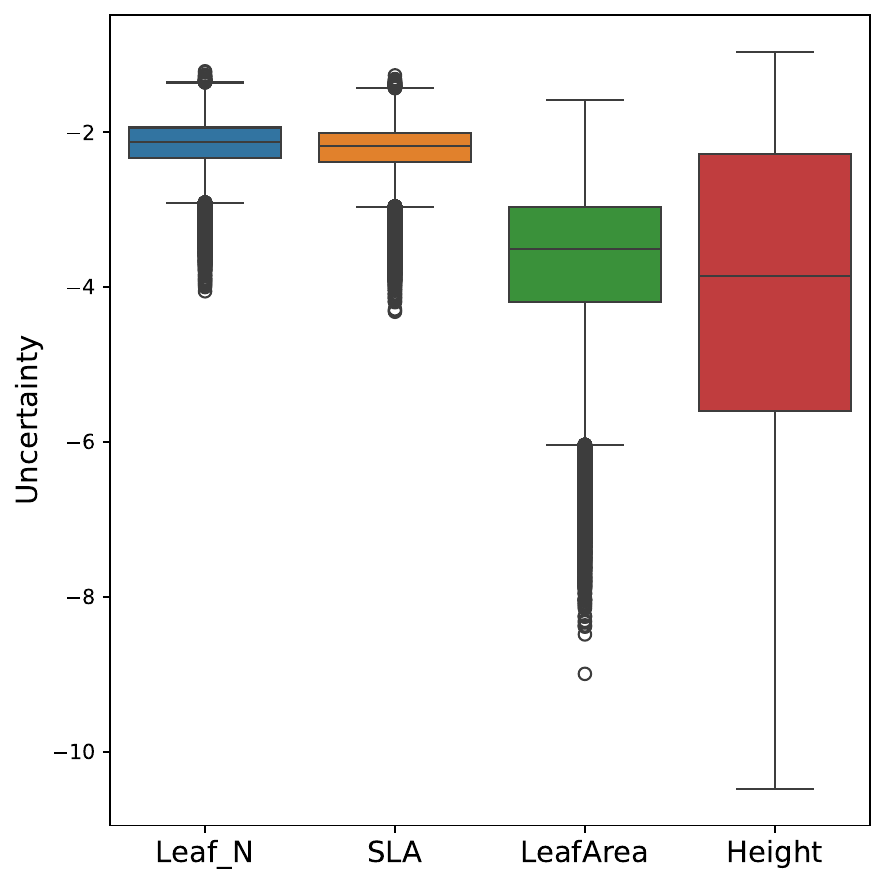}
\caption{Visualization of the predicted log-variance values across all traits during the uncertainty-aware filtering process. Here, Leaf\_N denotes leaf nitrogen, and SLA refers to specific leaf area.}
\label{fig:logvar_vis}
\end{figure}

For Height (H), Specific Leaf Area (SLA), and Leaf Nitrogen (LN), we assume a Gaussian distribution with predicted mean \( \hat{\mu}_n^m \) and variance \( \sigma_n^{2,m} = \exp(\hat{s}_n^m) \), where \( \hat{s}_n^m \) is the predicted log-variance. The negative log-likelihood (NLL) is used as the training objective: for LA, the Laplace NLL is
\[
\mathcal{L}_{\text{NLL}}^{m} = \frac{1}{N} \sum_{n=1}^{N} \left[ \frac{\left| y_n^m - \hat{\mu}_n^m \right|}{\exp(\hat{s}_n^m)} + \hat{s}_n^m \right],
\]
and for the remaining traits, the Gaussian NLL is
\[
\mathcal{L}_{\text{NLL}}^{m} = \frac{1}{N} \sum_{n=1}^{N} \left[ \frac{(y_n^m - \hat{\mu}_n^m)^2}{2 \exp(\hat{s}_n^m)} + \frac{1}{2} \hat{s}_n^m \right].
\]
This formulation enables the model to learn both the central tendency and the predictive uncertainty for each trait. We visualize predicted log-variance across traits in Fig. \ref{fig:logvar_vis}

\begin{figure*}[ht]
\centering
\includegraphics[width=\linewidth]{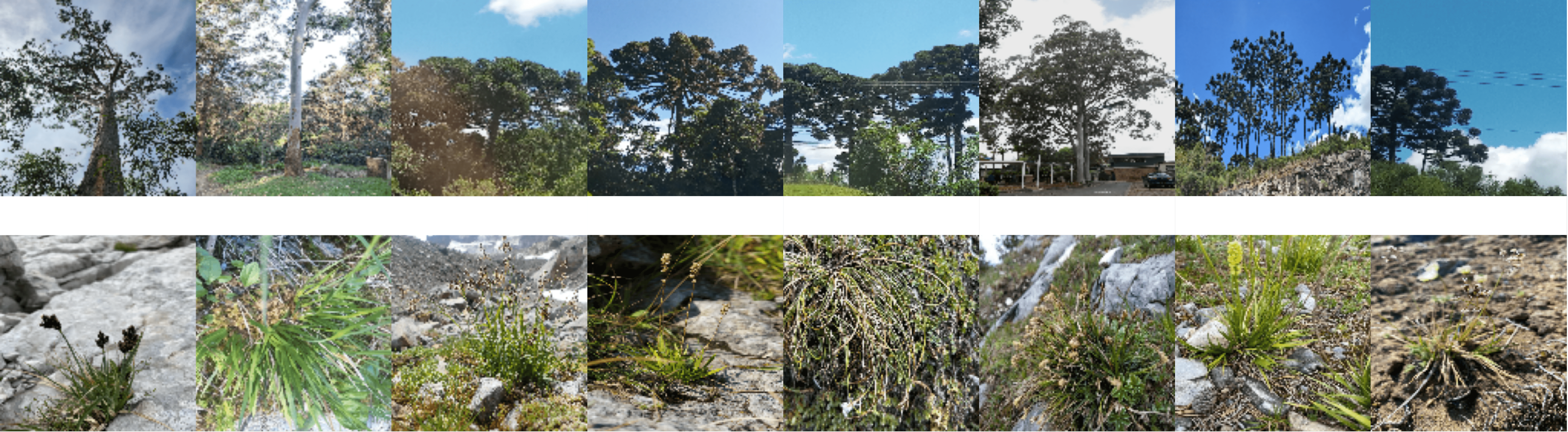}
   \caption{
Visualizations of predictive uncertainty for plant height during residual-aware filtering. Example images with high uncertainty (top) and low uncertainty (bottom) are shown. 
}
\label{fig:heteroscedasticityheight}
\end{figure*}

\begin{figure*}[ht]
  \centering
  % First row
  \begin{subfigure}[b]{\textwidth}
    \includegraphics[width=\linewidth]{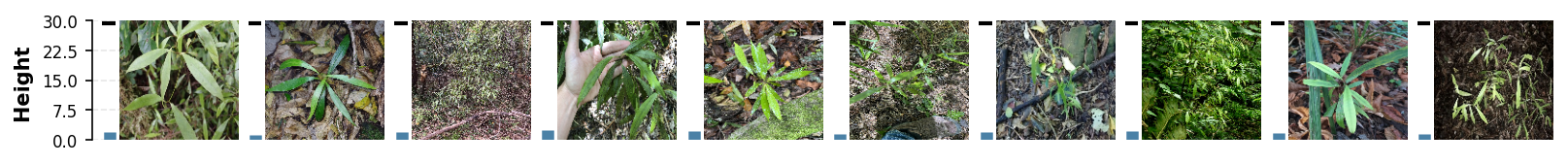}
    %\caption{Height}
  \end{subfigure}
  \hfill
  \begin{subfigure}[b]{\textwidth}
    \includegraphics[width=\linewidth]{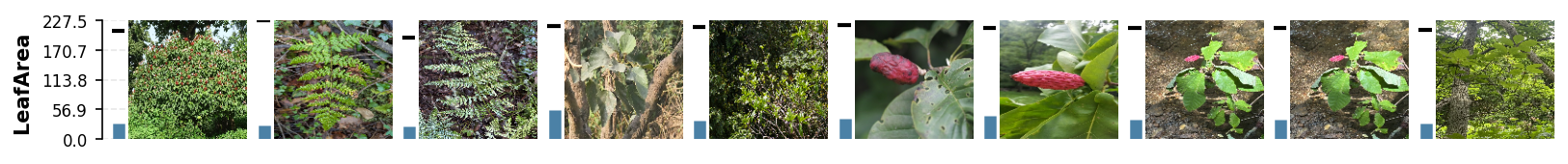}
    %\caption{LA}
  \end{subfigure}
  \hfill
   \caption{Examples of images with high uncertainty and high residual error identified during residual-aware filtering for Height (m) (top) and Leaf Area (cm\textsuperscript{2}) (bottom). Predicted trait values are shown as bars, with species median values indicated by black dashed lines. For Height, juvenile or undeveloped individuals exhibit low predicted height but high residual error due to elevated species-level medians, leading to their removal. For Leaf Area, high residuals are associated with exotic ferns, images not focused on leaves, and small leaves from species with large median leaf areas.}
  \label{fig:residualerror}
\end{figure*}

\subsection{Uncertainty-Guided Data Cleaning Loop}
\label{sec:uncerappendix}
To enable scalable data curation in the presence of label noise, we propose a two-step data cleaning strategy guided by model-predicted uncertainty and residual error.

\subsubsection*{Stage 1: Uncertainty-Aware Filtering}
We employ an iterative, uncertainty-aware filtering strategy to clean the training data while a model is being trained.

The model is first trained for one epoch on the full dataset. After this initial pass, we perform inference on the training set and compute predictive uncertainty across all four trait heads for each image. We then remove samples where the joint uncertainty falls within the top 5\%, as we hypothesize these are likely to be visually ambiguous or noisy. Examples of such samples include poorly lit or occluded images, or those containing non-foliage content like twigs or fruit.

The model's training then continues from its current state on this filtered dataset. This filtering–retraining cycle is repeated for up to two iterations or until the number of high-uncertainty samples drops below a predefined threshold.

\subsubsection*{Step 2: Residual-Aware Filtering}
As visualized in \Cref{fig:heteroscedasticityheight}, we observe that predictive uncertainty for traits like height increases for taller individuals, a phenomenon we hypothesize is due to heteroscedasticity. This is consistent with our observation that residuals for taller plants are generally larger than for smaller plants, leading the model to output a correspondingly higher uncertainty for taller plants. Consequently, a naive filtering approach that relies solely on high predictive uncertainty could disproportionately remove samples from larger growth forms (e.g. trees) and fail to distinguish genuinely mislabeled data from correctly labeled but uncertain predictions.

To address this, we introduce residual-aware filtering. For each trait, we identify a `turning point', the training epoch after which performance on an external scientific reference set begins to decline, suggesting the onset of memorization. At this checkpoint, we compute normalized mean absolute error (nMAE) between predicted trait values and species-level medians. We hypothesize that samples exhibiting both high uncertainty (above the 95th percentile) and large residual error (above 50\%) are likely mislabeled or unreliable. These samples are removed over four iterations, each followed by training from scratch.

We visualize examples of such filtered samples for height and leaf area in \Cref{fig:residualerror}. For height, we often observe juvenile or undeveloped individuals with low predicted values but large residuals due to high species-level medians. For leaf area, examples for uncertainty samples are exotic ferns or unsharp photographs.

We continue the process until the number of samples with high-residual and high-uncertainty becomes negligible. In total, we remove approximately 500 species from an initial pool of 5,500. The resulting trait distributions are shown in \Cref{fig:traitdistribution}. While some borderline cases may be excluded, we argue that the volume of citizen science data compensates for a reduction of the total sample size and number of species. Furthermore, given the difficulty of verifying trait labels, particularly for biochemical traits such as specific leaf area (SLA) or leaf nitrogen, we prefer cautious removal over attempting label correction.

\subsection{Experiment}

\subsubsection{Experimental Setup}
\label{appendix:expsettings}

\textbf{Data Preprocessing.}
In both the global plant kingdom and citizen science datasets, grassland species are the most frequently observed growth form, followed by shrubs, with tree species being the least represented. Blindly training on such imbalanced data would likely bias the model toward frequent grassland species, reducing its ability to generalize to less common growth forms. To mitigate this, we stratified the data by growth forms and applied weighted sampling during training. Each batch was constructed to contain a balanced mix of grasses, shrubs, and trees, ensuring more equitable learning across growth forms. Lastly, all continuous target traits were normalized to the range [0, 1] using MinMax scaling.
%Each training batch contains a balanced representation of growth forms to prevent dominance by overrepresented species or forms.

\textbf{Training Procedure.} The model is trained for 30 epochs using the AdamW optimizer with $\beta=(0.9, 0.999)$, and a weight decay of $5 \times 10^{-5}$. A cosine annealing learning rate schedule decays the learning rate from an initial value of $1 \times 10^{-5}$ to a minimum of $5 \times 10^{-6}$. Training stability is further ensured by gradient clipping with a maximum norm of 1.0.

\subsection{Results}

\subsubsection{Global Maps Trait Maps derived from Aggregated PlantTraitNet Predictions}

\begin{figure}[hp]
\centering
\includegraphics[width=0.5\textwidth]{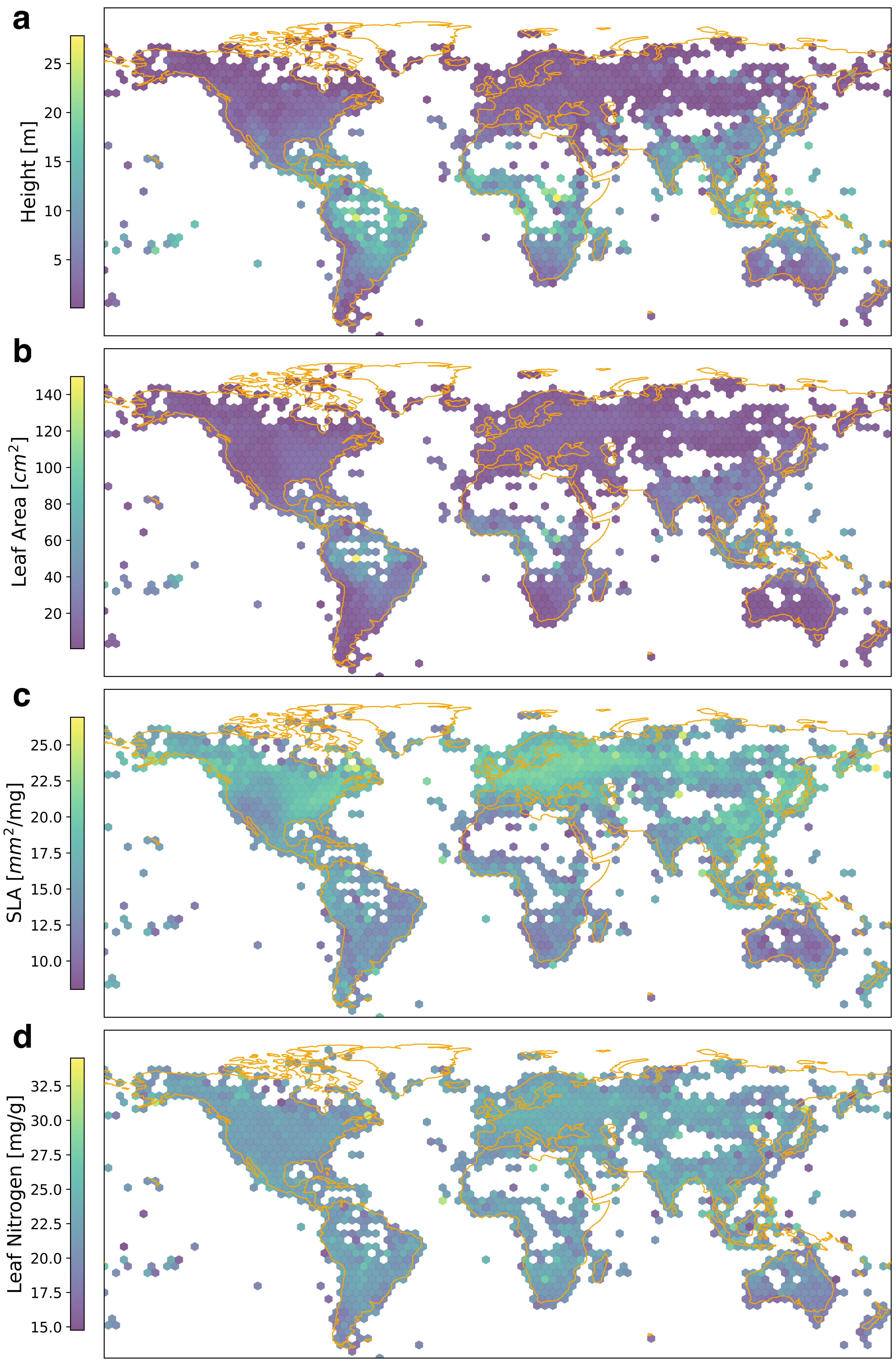}
\caption{Global trait maps derived from spatially aggregating PlantTraitNet predictions at 1 degree resolution. The global trait maps comprise height (a), SLA (b), leaf area (c), and leaf nitrogen (d) based on the spatial aggregation of citizen science photos from the validation dataset.}
\label{fig:prediction-maps}
\end{figure}

\Cref{fig:prediction-maps} show global maps at 1 degree resolution obtained from aggregating predicted traits from PlantTraitNet on 300k citizen science images. In addition to the quantitative evaluation against vegetation surveys (sPlotOpen, next section), we found that the global patterns reproduce expected global trait patterns \citep{lusk2025smartphones, wolf2022citizen}. For instance, for height and leaf area, we find expected longitudinal variation, such as large plants with big leaves towards the equator. Most robust plants (high specific leaf area and nitrogen content) are found in temperate and boreal zones.

\subsubsection{Evaluating Global Trait Maps from PlantTraitNet with Vegetation Survey data (sPlotOpen)}
To empirically demonstrate the effectiveness of our uncertainty-guided data cleaning process, we compare model performance on raw (unfiltered) and refined dataset (Table~\ref{tab:suppl_globalbenchmark}). Performance is evaluated using $R^2$, nMAE, and Pearson’s $r$, with results averaged over three independent runs.

We observe that our data refinement process leads to consistent improvements in predictive performance. Notably, $R^2$ for SLA and LA increases from 0.23 to 0.27 and 0.30 to 0.34, respectively, demonstrating enhanced predictive reliability. For Leaf Nitrogen (LN), although the model's overall performance is limited (negative $R^2$), the refined dataset still yields a marginal improvement in correlation ($r = 0.50$ vs. 0.49).

\begin{figure*}[t] % 'h' means place the figure approximately here
    \centering
    \includegraphics[width=0.99\textwidth]{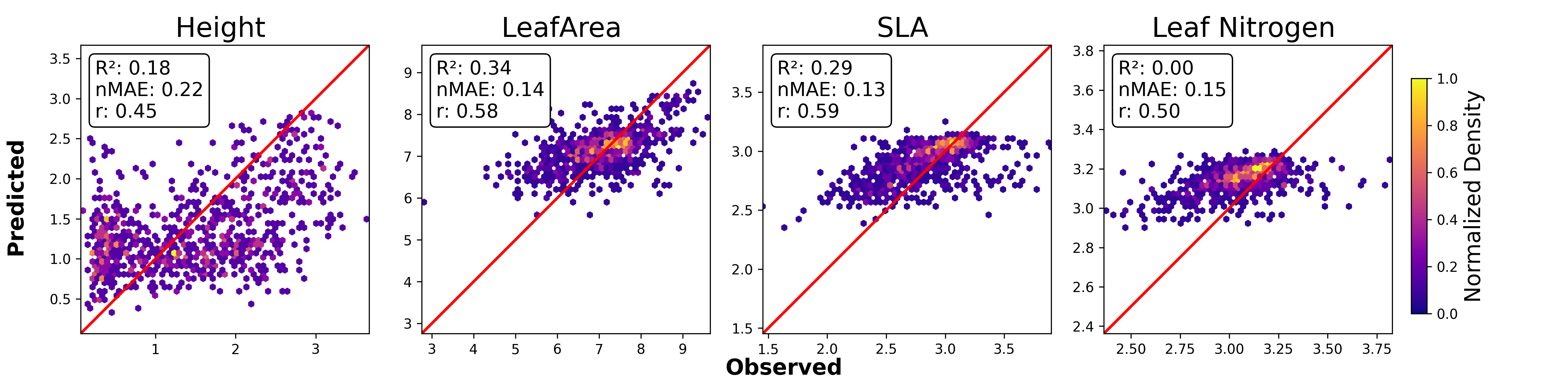} 
    % Replace with your image file name and extension
    \caption{Global trait predictions obtained from PlantTraitNet against globally distributed vegetation survey data (sPlotOpen). Trait predictions of geotagged images are aggregated at a 1° spatial resolution using image geolocations (Predicted). These aggregated predictions are then compared with averaged community-weighted trait values derived from sPlotOpen (Observed).}
    \label{fig:scatterplot}
\end{figure*}

The results, visualized in \Cref{fig:scatterplot}, indicate that filtering noisy samples is a key step toward more robust and reliable trait predictions from our citizen science data.

\begin{table*}
\caption{
Comparison of PlantTraitNet with sPlotOpen community weighted trait means at 1\textdegree{} resolution. We report mean $\pm$ standard deviation across 3 runs with different random initializations. Raw refers to models trained with unfiltered citizen science data, while refined refers models based on the uncertainty-aware training.
}

\label{tab:suppl_globalbenchmark}
\centering
\begin{tabular}{l c c c c c}
\toprule
Method & Metric & H & LA & SLA & LN \\
\midrule
\multirow{4}{*}{Ours (Raw)} 
& $R^2$ ↑   & \textbf{0.19 ± 0.01} & 0.30 ± 0.04 & 0.23 ± 0.01 & -0.16 ± 0.03
% & \multirow{3}{*}{$\sim890$} 
\\
& nMAE ↓    & \textbf{0.22 ± 0.00} & \textbf{0.14 ± 0.00} & 0.14 ± 0.00 & \textbf{0.17 ± 0.00} \\ 
& $r$ ↑     & \textbf{0.45 ± 0.01} & \textit{0.56 ± 0.01} & \textbf{0.59 ± 0.01} & \textit{0.49 ± 0.01} \\ 
\midrule
\multirow{4}{*}{Ours (Refined)} 
& $R^2$ ↑   & \textit{0.18 ± 0.00} & \textbf{0.34 ± 0.01} & \textbf{0.27 ± 0.02} & \textbf{-0.12 ± 0.14} \\
%& \multirow{3}{*}{$\sim890$}\\ 
& nMAE ↓    & \textbf{0.22 ± 0.0} & \textbf{0.14 ± 0.0} & \textbf{0.13 ± 0.00} & \textbf{0.17 ± 0.01} \\ 
& $r$ ↑     & \textbf{0.45 ± 0.00} & \textbf{0.57 ± 0.00} & \textbf{0.59 ± 0.01} & \textbf{0.50 ± 0.00} \\ 
%&  \#grids  & 899 & 896 & 901 & 891  \\ 

\bottomrule
\end{tabular}
\end{table*}

\subsubsection{Qualitative Assessment of Intraspecific Trait Variability}

Trait values within a species can vary across environmental gradients and hence an accurate representation of intraspecific variability can be important to create global trait maps.
However, the weak annotations used for model training were based on species-level matching of citizen science photographs and trait values from the TRY database.
Accordingly, the model was not explicitly trained to capture within-species variability.
To qualitatively evaluate if PlantTraitNet still captures intraspecific trait variation, we analyzed model predictions for ecologically diverse species across major growth forms, including grasses, shrubs, and trees. For each selected species, we conducted two complementary analyses:

\begin{enumerate}
\item We visualized predicted trait values for a small, held-out subset of seven individuals that exhibited notable visual variation in developmental stage, size or structure.
\item We compared the distribution of predicted trait values from up to 100 training images to the corresponding distribution of observed trait values from up to 100 samples in the TRY database.
\end{enumerate}

\begin{figure*}[htbp]
  \centering
  \includegraphics[width=\linewidth]{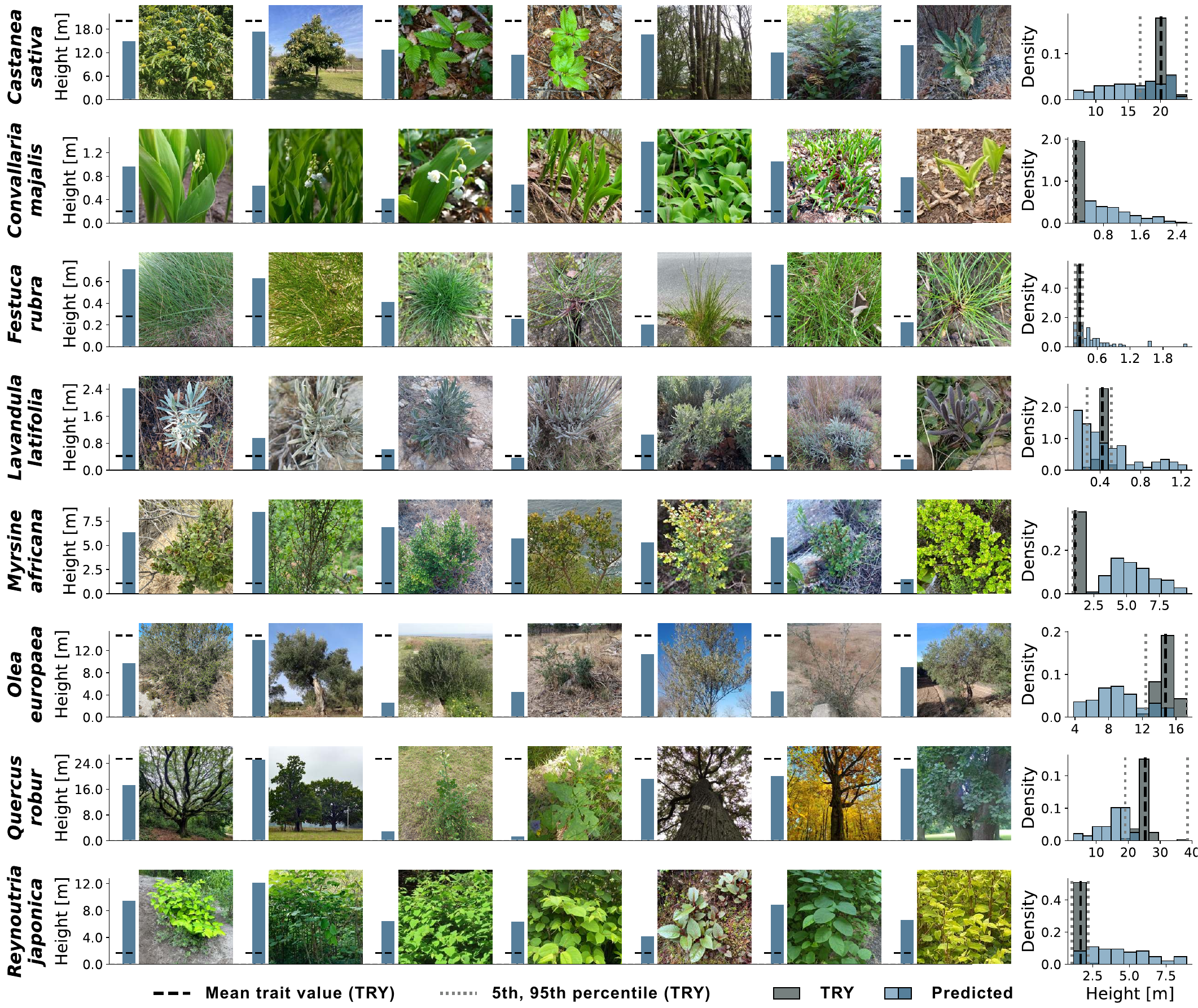}
  \caption{Intraspecific variation in predicted plant height compared to TRY-derived trait means.}
  \label{fig:quali-h}
\end{figure*}

\begin{figure*}[htbp]
  \centering
  \includegraphics[width=\linewidth]{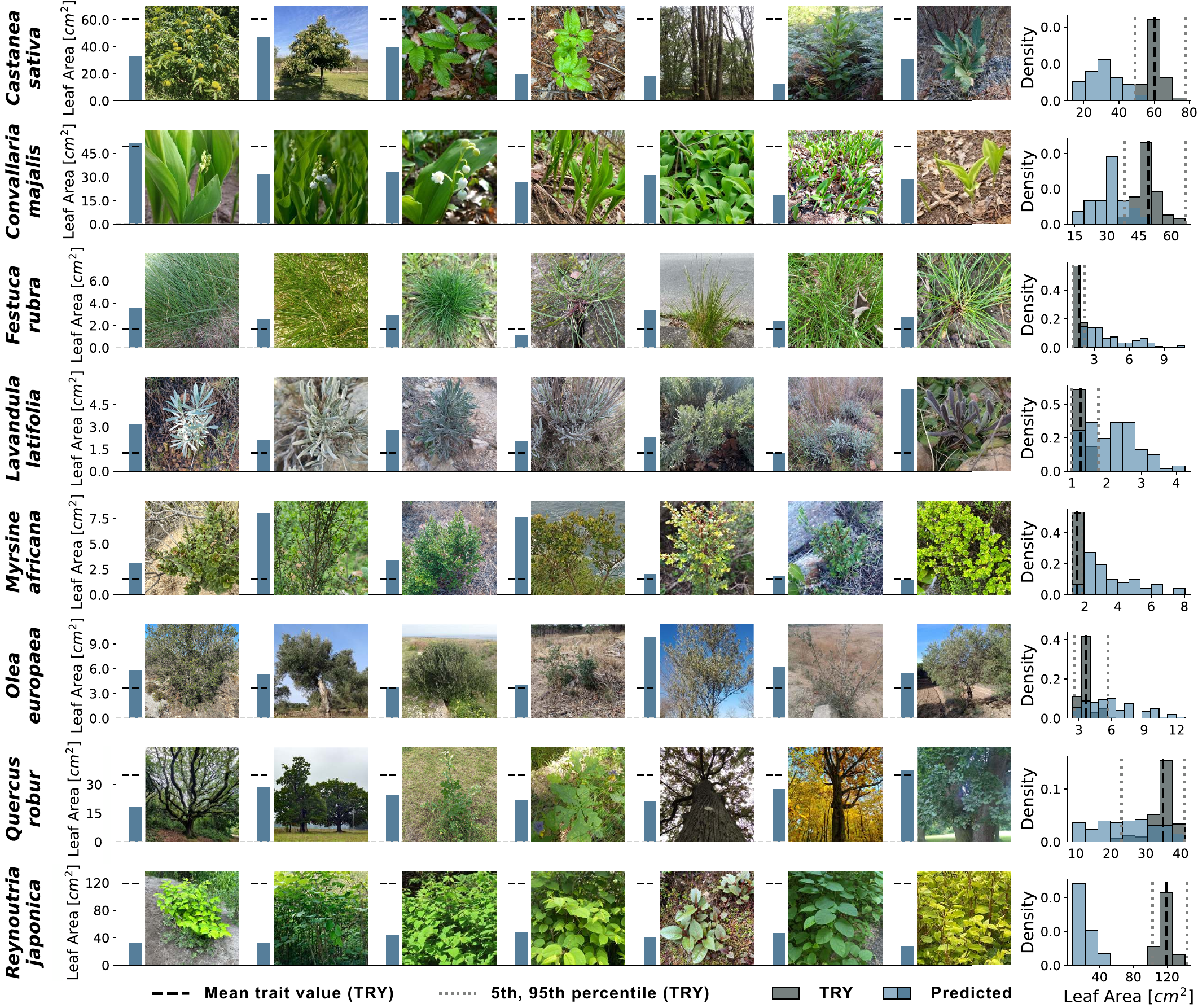}
  \caption{Intraspecific variation in predicted leaf area compared to TRY-derived trait means.}
  \label{fig:quali-la}
\end{figure*}

\begin{figure*}[htbp]
  \centering
  \includegraphics[width=\linewidth]{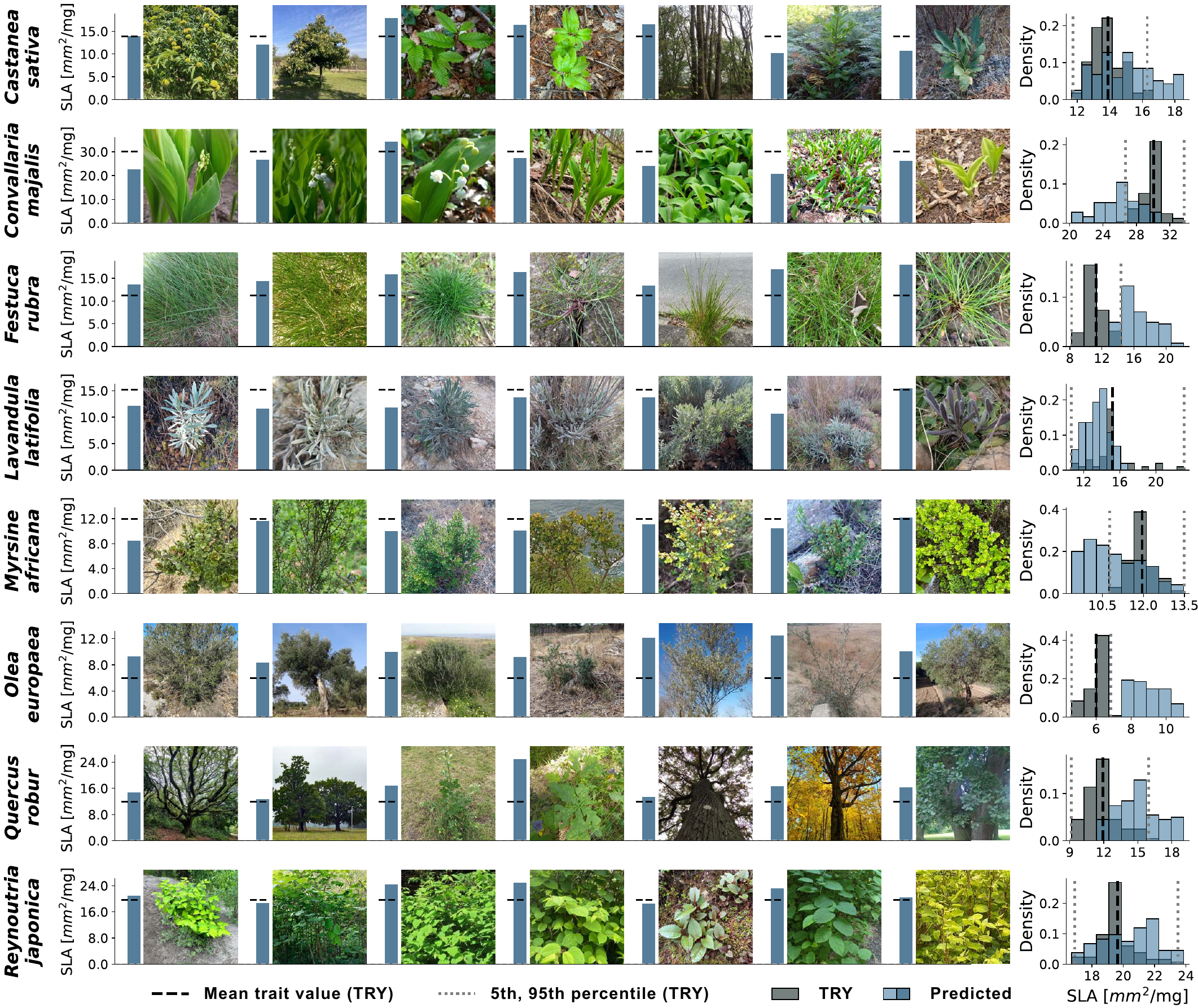}
  \caption{Intraspecific variation in predicted specific leaf area compared to TRY-derived trait means.}
  \label{fig:quali-sla}
\end{figure*}

\begin{figure*}[htbp]
  \centering
  \includegraphics[width=\linewidth]{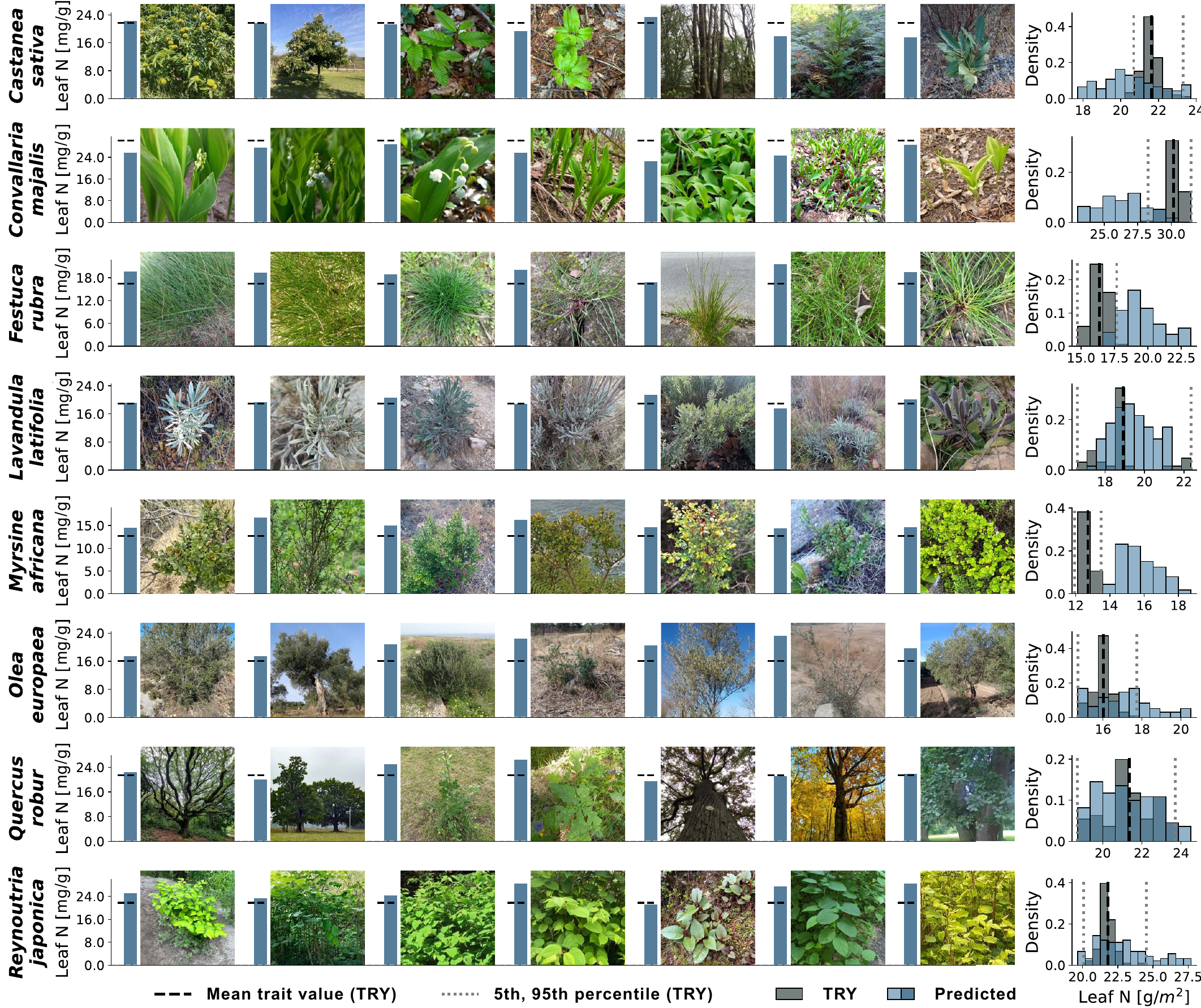}
  \caption{Intraspecific variation in predicted leaf nitrogen compared to TRY-derived trait means.}
  \label{fig:quali-ln}
\end{figure*}

This analysis provides insight into how well the model reflects trait variability within species, relative to aggregated trait observations.
Our results, visualized in \Cref{fig:quali-h,fig:quali-la,fig:quali-sla,fig:quali-ln}, consistently show that the model learns to represent a wide range of intraspecific variation for traits like Height (H) and Leaf Area (LA). In these cases, the predicted trait ranges often exceeded the variation captured in the original TRY database observations. This seems plausible, given that trait observations are typically performed on healthy and adult plants.
Compared to Leaf Area (LA) and Plant Height (H), for physiological traits such as Specific Leaf Area (SLA) and Leaf Nitrogen (LN), we observe little predicted variation. We hypothesize this is a positive outcome, as these leaf traits are ecologically expected to show less intraspecific variability on the premise that interspecific trait variation generally exceeds intraspecific variation \cite{dong2020components, wright2017global}. For improved visibility, only predicted values between the 5th and 95th percentiles are shown.

\begin{table}[htbp]
\centering
\renewcommand{\arraystretch}{1.2}
\caption{Phylogenetic signal metrics with corresponding p-values}
\begin{tabular}{lcccc}
\hline
\textbf{Trait Error} & \textbf{K} & \textbf{K*} & \textbf{Lambda} \\
\hline
\multirow{2}{*}{H} 
& 0.018 & 0.015 & 0.801 \\
& {\footnotesize (p = 0.001)} & {\footnotesize (p = 0.001)} & {\footnotesize (p = 0.001)} \\
\hline
\multirow{2}{*}{LA} 
& 0.007 & 0.007 & 0.556 \\
& {\footnotesize (p = 0.064)} & {\footnotesize (p = 0.079)} & {\footnotesize (p = 0.001)} \\
\hline
\multirow{2}{*}{SLA} 
& 0.005 & 0.006 & 0.042 \\
& {\footnotesize (p = 0.441)} & {\footnotesize (p = 0.399)} & {\footnotesize (p = 0.003)} \\
\hline
\multirow{2}{*}{LN} 
& 0.008 & 0.008 & 0.150 \\
& {\footnotesize (p = 0.001)} & {\footnotesize (p = 0.001)} & {\footnotesize (p = 0.001)} \\
\hline
\end{tabular}
\label{tab:phylo_signal}
\end{table}

\subsubsection{Trait Prediction Error Along the Taxonomic Tree}

\begin{figure}[htpb]
\centering
\includegraphics[width=0.47\textwidth]{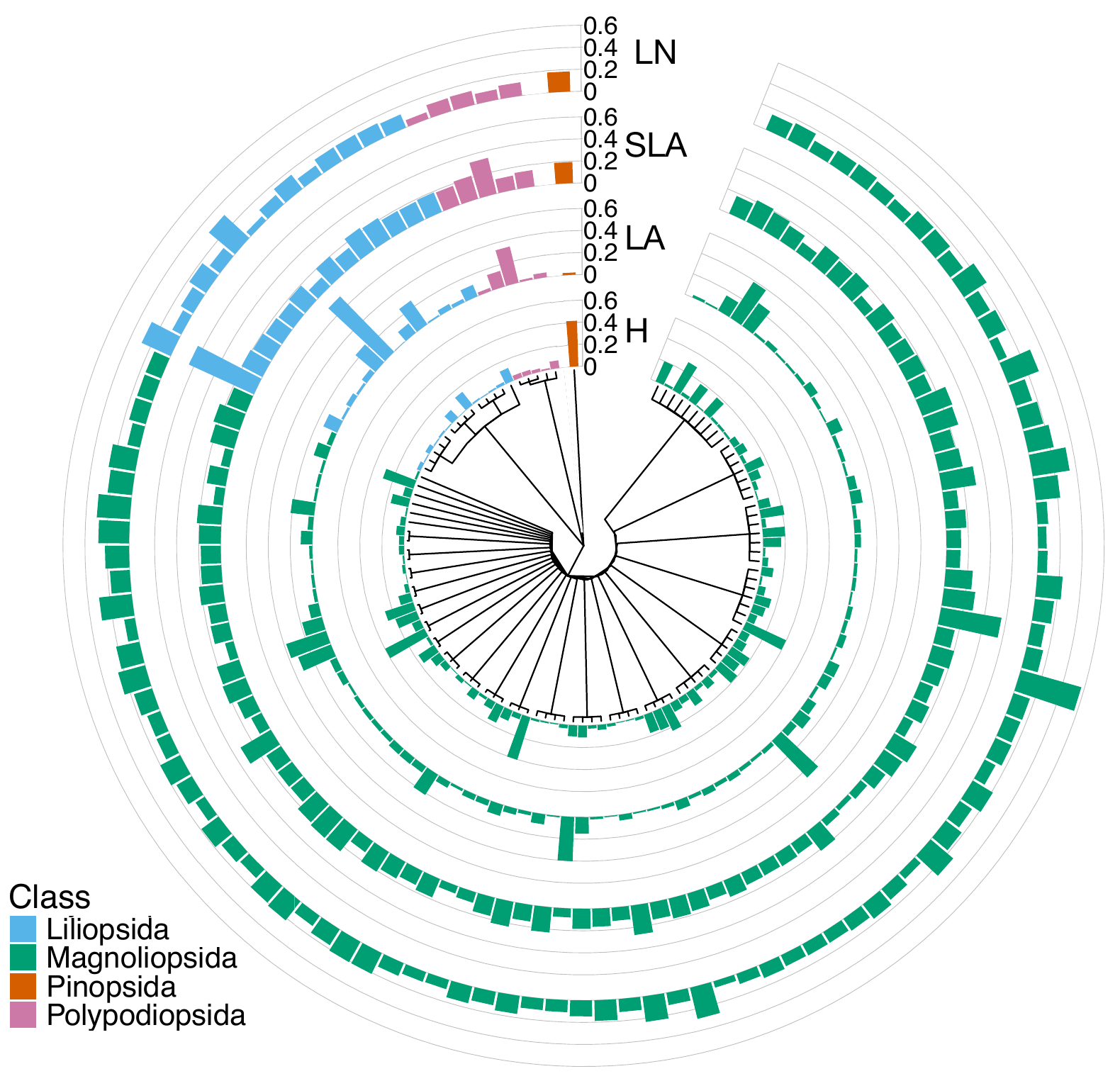}
\caption{Mean relative prediction error (MRPE) computed on validation data at the family level, visualized along the taxonomic tree, for height (H), leaf area (LA), specific leaf area (SLA) and leaf nitrogen (LN).}
\label{fig:phylo_family}
\end{figure}

A phylogenetic tree was constructed from hierarchical taxonomic information (Kingdom to Species) using as.phylo() from the ape package \citep{paradis2019ape}. Mean relative errors (MRE) were computed per species, normalized by the trait-specific error range, and matched with taxonomic metadata. In \Cref{fig:phylo_family}, MREs are shown at the family and class levels, with no clear visual pattern of systematic bias. 
Moreover, we report standard metrics to quantify the phylogenetic signal in species-level prediction errors, calculated via the phyloSignal() function from the phylosignal R package. Pagel’s $\boldsymbol{\lambda}$ \citep{pagel1999inferring} measures the overall fit of the data to the phylogeny by scaling internal branches; values close to 1 indicate a strong phylogenetic structure, while values near 0 suggest independence from phylogeny. Statistical significance is assessed by comparing the fitted $\lambda$ to a model with $\lambda = 0$. Blomberg’s $\boldsymbol{K}$ \citep{blomberg2003testing} compares the observed trait variance across the tree to expectations under a Brownian motion model; values close to 1 indicate strong signal, and values $\ll$1 suggest weak signal, especially among closely related species. $\boldsymbol{K^*}$ is a variation of $K$ that is less sensitive to tree imbalance and branch length distortions. Together, these metrics offer complementary perspectives on phylogenetic structure at different evolutionary scales (see \Cref{tab:phylo_signal}).
Pagel’s $\lambda$ captures broad-scale phylogenetic autocorrelation in residual covariance \citep{pagel1999inferring}, while Blomberg’s $K$ is more sensitive to fine-scale signal among closely related species \citep{blomberg2003testing}. For SLA and leaf nitrogen, the phylogenetic signal is weak ($\lambda$ = 0.04 and 0.15; $K$ = 0.0053 and 0.0076), suggesting that prediction errors are largely independent of species relatedness. Although errors for height ($\lambda$ = 0.80, $K$ = 0.018) and leaf area ($\lambda$ = 0.56, $K$ = 0.0067) show some phylogenetic autocorrelation, the consistently low $K$ values indicate that even closely related species do not share systematic prediction biases. These results reveal that while slight patterns may emerge at the level of broader plant classes, there is no consistent phylogenetic structure among closely related species. Altogether, this underscores that despite being trained on species-level annotations, the model exhibits strong generalizability and methodological robustness for predicting plant traits across the plant kingdom.

\begin{table*}[htbp]
  \centering
  \caption{
  Impact of trait-specific loss functions on predictive performance. Reported values are mean $R^2$ ± one standard deviation across three independently initialized training runs for each loss configuration.
  }
  \label{tab:loss_ablation}
  \begin{tabular}{llcccc}
    \toprule
    Method & H & LA & SLA & LN \\
    \midrule
    Gaussian &  0.18 ± 0.02 & 0.31 ± 0.01 & 0.31 ± 0.01 & \textbf{0.18 ± 0.01} \\
    Laplace (LA + H) & 0.10 ± 0.01 & \textbf{0.32 ± 0.01} & \textbf{0.32 ± 0.02} & \textbf{0.18 ± 0.03} \\
    Laplace (LA) & \textbf{0.19 ± 0.02} & \textbf{0.32 ± 0.01} & \textit{0.31 ± 0.02} & \textbf{0.18 ± 0.05} \\
    \bottomrule
  \end{tabular}
\end{table*}
\begin{table*}[htbp]
\centering
\caption{
Integrating embeddings from patch token vs. classification token (CLS) in image encoders. We report $R^2 \pm$ 1 standard deviation across 3 runs.
}
\label{tab:clspatchtoken}

\begin{tabular}{lcccc}
\hline
Image Encoder & H & LA & SLA & LN \\
\hline
DINOv2 (CLS token)   & \textbf{0.18 ± 0.01} & \textbf{0.31 ± 0.00} & 0.30 ± 0.01 & 0.11 ± 0.01 \\
DINOv2 (patch mean)  & 0.15 ± 0.00 & \textbf{0.31 ± 0.00} & \textbf{0.32 ± 0.00} & \textbf{0.14 ± 0.01} \\
\hline
BioCLIP (CLS token)  & 0.15 ± 0.01 & \textbf{0.30 ± 0.00} & \textbf{0.32 ± 0.01} & \textbf{0.15 ± 0.04} \\
BioCLIP (patch mean) & \textbf{0.17 ± 0.01} & \textbf{0.30 ± 0.01} & 0.31 ± 0.01 & 0.12 ± 0.05 \\
\hline
\end{tabular}
\end{table*}

\begin{table*}[htp]
  \caption{Effect of embedding size for each modality and the multimodal backbone. We perform ablations over different embedding dimensions: [512, 768] for image encoders, [256, 512, 1024] for geospatial encoders, and [768, 1024] for the multimodal fusion layer. We report $R^2 \pm$ 1 standard deviation across 3 runs.
}
  \label{tab:ablation_embeddingsize}
  \centering
  \begin{tabular}{l l c c c cc c c}
    \hline
    \multirow{2}{*}{Image} & \multirow{2}{*}{Geo} & Multimodal &  \multirow{2}{*}{H}& \multirow{2}{*}{LA}& \multirow{2}{*}{SLA}&\multirow{2}{*}{LN}\\
    &   &  Backbone Dim & & &&  \\
    \hline
     \multicolumn{7}{l}{DinoV2(patch) + Climplicit}\\
    \hline
    \multirow{1}{*}{512} & \multirow{1}{*}{256} & \multirow{1}{*}{768}& 0.18 ± 0.01 & 0.30 ± 0.01 & 0.30 ± 0.01 & 0.13 ± 0.04 \\
    \hline
    \multirow{1}{*}{768} & \multirow{1}{*}{256} & \multirow{1}{*}{768}  & \textbf{0.19 ± 0.01} & \textbf{0.32 ± 0.01} & \textbf{0.31 ± 0.01} & \textbf{0.16 ± 0.06} \\
    \hline
    \multirow{1}{*}{768} & \multirow{1}{*}{256} & \multirow{1}{*}{1024}& 0.18 ± 0.02 & 0.30 ± 0.03 & 0.29 ± 0.01 & 0.12 ± 0.02 \\
    \hline

    \multirow{1}{*}{768} & \multirow{1}{*}{1024} & \multirow{1}{*}{768} & 0.18 ± 0.00 & 0.31 ± 0.01 & 0.29 ± 0.02 & 0.12 ± 0.03 \\
    \hline
    \multirow{1}{*}{768} & \multirow{1}{*}{1024} & \multirow{1}{*}{1024}& 0.17 ± 0.01 & 0.30 ± 0.03 & 0.28 ± 0.01 & 0.13 ± 0.03 \\
    \hline
    \multirow{1}{*}{768} & \multirow{1}{*}{512} & \multirow{1}{*}{768} 
    & 0.18 ± 0.01 & \textbf{0.32 ± 0.02} & 0.29 ± 0.01 & 0.14 ± 0.05 \\

    \hline
    \multirow{1}{*}{768} & \multirow{1}{*}{512} & \multirow{1}{*}{1024}  
    %R\textsuperscript{2} 
    & 0.17 ± 0.01 & 0.31 ± 0.01 & 0.3 ± 0.01 & 0.14 ± 0.02 \\
    \hline

    \multicolumn{7}{l}{DinoV2(patch) + SatCLIP}\\
    \hline
     \multirow{1}{*}{768} & \multirow{1}{*}{256} & \multirow{1}{*}{768} 
    & \textbf{0.16 ± 0.02} & \textbf{0.27 ± 0.04} & \textbf{0.25 ± 0.02} & 0.11 ± 0.05 \\

     \hline
     \multirow{1}{*}{768} & \multirow{1}{*}{256} & \multirow{1}{*}{1024} 
& 0.13 ± 0.03 & 0.26 ± 0.03 & \textbf{0.25 ± 0.01} & \textbf{0.13 ± 0.01} \\
    \hline

    \multicolumn{7}{l}{DinoV2(patch) + GeoCLIP}\\
    \hline
         \multirow{1}{*}{768} & \multirow{1}{*}{256} & \multirow{1}{*}{768} 
      & \textbf{0.17 ± 0.01} & \textbf{0.33 ± 0.01} & \textbf{0.32 ± 0.00} & \textbf{0.15 ± 0.01} \\
     \hline
     \multirow{1}{*}{768} & \multirow{1}{*}{256} & \multirow{1}{*}{1024} 
     & 0.16 ± 0.01 & \textbf{0.33 ± 0.00} & 0.31 ± 0.01 & \textbf{0.15 ± 0.02} \\

     \hline
     \multirow{1}{*}{768} & \multirow{1}{*}{512} & \multirow{1}{*}{768} 
     & \textbf{0.17 ± 0.0} & 0.32 ± 0.01 & \textbf{0.32 ± 0.00} & 0.13 ± 0.02 \\
     \hline
    \multirow{1}{*}{768} & \multirow{1}{*}{512} & \multirow{1}{*}{1024}
    & \textbf{0.17 ± 0.0} & 0.3 ± 0.0 & 0.31 ± 0.01 & 0.13 ± 0.02 \\
    \hline
    \multicolumn{7}{l}{DinoV2(patch) + Climplicit(256-D) + Depth(768-D)}\\
    \hline
     \multirow{1}{*}{768} & \multirow{1}{*}{256} & \multirow{1}{*}{768} 
& 0.17 ± 0.02 & \textbf{0.32 ± 0.02} & \textbf{0.32 ± 0.01} & 0.16 ± 0.03 \\

     \hline
     \multirow{1}{*}{768} & \multirow{1}{*}{256} & \multirow{1}{*}{1024} 
    & \textbf{0.19 ± 0.02} & \textbf{0.32 ± 0.01} & \textit{0.31 ± 0.02} & \textbf{0.18 ± 0.05} \\

    \bottomrule
  \end{tabular}
\end{table*}

\section{Ablation Study}
\subsection{Effect of Trait-Specific Loss Functions}

Plant traits can differ strongly in their distributions, which in turn can affect the model training.
To investigate the influence of loss function choice on trait prediction, we compare Gaussian and trait-specific Laplace loss formulations, as shown in Table~\ref{tab:loss_ablation}. Given the long-tailed distributions observed for certain traits, notably leaf area (LA) and height (H) (see \Cref{fig:traitdistribution}), we hypothesize that the heavier-tailed Laplace loss may improve model robustness and calibration.
Applying the Laplace loss to LA results in performance improvements compared to the Gaussian loss for both LA and height, while performance for SLA and LN remains similar. However, applying the Laplace loss to both LA and H decreases performance for height. This outcome may be related to the height distribution being heavily influenced by grass species, causing imbalance. Since we mitigate this imbalance through stratified sampling across plant functional types such as grasses, shrubs, and trees, the Gaussian loss appears more appropriate for modeling height under these conditions.
In summary, our results suggest that trait-specific loss functions can benefit LA prediction. Based on this analysis, we adopt a hybrid approach in the final model, using the Laplace loss for LA and Gaussian losses for the other traits. This approach is associated with improved predictive performance across traits and underlines that model training requires careful consideration of multivariate ecological complexity.

\subsubsection{Effect of Token Pooling Strategy on Trait Prediction Performance }
While the classification token (`[CLS]') is a common choice for a global image representation, the individual patch tokens of vision transformers contain rich, spatially-aware information that can be more beneficial for downstream tasks \cite{caron2021emerging}. We therefore perform an ablation study to determine the optimal token representation for our trait prediction task (Table~\ref{tab:clspatchtoken}).

 We observe that for DinoV2 patch token pooling improves $R^2$ scores for SLA (from 0.30 to 0.32) and LN (from 0.11 to 0.14), while Height performs better with the classification token ($R^2$ of 0.18 vs. 0.15), and Leaf Area shows no change. Based on the net gains for SLA and LN, we use patch token pooling in our final configuration.

For BioCLIP, we conduct a similar ablation and observe that the classification token consistently performs better on average. We hypothesize that this may be due to the contrastive training objective of BioCLIP, which aligns the classification token with the projected textual representation in a joint embedding space, making it a more robust representation for this model.

\subsubsection{Effect of Embedding Dimensionality on Multimodal Trait Prediction}
The choice of embedding dimensionality and model capacity is a critical design decision that balances computational efficiency with representational power. To find the optimal configuration for our model, we conducted a detailed ablation study to systematically evaluate the impact of embedding dimensions across our primary modalities and the multimodal fusion backbone (Table~\ref{tab:ablation_embeddingsize}).
 
We vary (i) the image embedding dimension (512 vs. 768), (ii) the geolocation embedding dimension (256, 512, and 1024), and (iii) the hidden size of the multimodal fusion backbone (768 vs. 1024).

We observe that using a 768-dimensional image embedding from DINOv2, a 256-dimensional geolocation embedding, and a 768-dimensional multimodal fusion backbone tends to yield strong performance across traits. This trend is consistent across the tested geolocation encoders: Climplicit~\cite{dollinger2025climplicit}, SatCLIP~\cite{klemmer2025satclip}, and GeoCLIP~\cite{vivanco2023geoclip}. Among these, Climplicit performs best in our experimental setup.

When including depth as an additional modality, performance improves when the multimodal backbone is increased to 1024 dimensions. We hypothesize that this benefit may arise from the increased representational capacity required to integrate the added modality. Based on these findings, we use a 768-dimensional multimodal backbone when the model includes only image and geolocation inputs, and a 1024-dimensional backbone when depth is incorporated.